\documentclass[a4paper,fleqn]{cas-dc}

\usepackage[numbers]{natbib}
\usepackage{xcolor}
\usepackage{kotex}
\usepackage{booktabs}
\usepackage{float}
\usepackage{xurl}
\usepackage{makecell}
\usepackage{subfig}
\usepackage[breakable]{tcolorbox}
\usepackage{subfig}
\usepackage[dvipsnames]{xcolor}
\usepackage{wasysym}
\usepackage{algorithm}
\usepackage{algpseudocode}

\newcommand{\ourtool}{\textit{HalluVision}}

\begin{document}
\let\WriteBookmarks\relax
\def\floatpagepagefraction{1}
\def\textpagefraction{.001}

\shorttitle{Measuring Hallucination in Large Language Models for Cyber Threat Intelligence: An Exploratory Study}    

\shortauthors{Song et al.}  

\title [mode = title]{Measuring Hallucination in Large Language Models for Cyber Threat Intelligence: An Exploratory Study}  

\author[korea]{Seungho Song}[orcid=0009-0003-6702-6723]
\ead{rickysong@korea.ac.kr}

\affiliation[korea]{organization={Korea University},
            addressline={145 Anam-ro, Seongbuk-gu}, 
            city={Seoul},
            postcode={02841}, 
            country={Republic of Korea}}

\affiliation[kwu]{organization={Kwangwoon University},
            addressline={20 Kwangwoon-ro, Nowon-gu}, 
            city={Seoul},
            postcode={01897}, 
            country={Republic of Korea}}

\affiliation[kaist]{organization={KAIST},
addressline={291 Daehak-ro, Yuseong-gu}, 
city={Daejeon},
postcode={34141}, 
country={Republic of Korea}}

\affiliation[cbnu]{organization={Chungbuk National University},
addressline={1 Chungdae-ro, Seowon-gu}, 
city={Cheongju, Chungcheongbuk-do},
postcode={28644}, 
country={Republic of Korea}}

\author[korea]{Donggyu Lee}[orcid=0009-0007-7682-5630]
\ead{tony7078@korea.ac.kr}

\author[kaist]{Jaehan Kim}[orcid=0000-0001-8048-097X]
\ead{jaehan@kaist.ac.kr}

\author[kwu]{Jiyoung Choi}[orcid=0009-0006-3314-0108]
\ead{jiyoungn01s@gmail.com}

\author[cbnu]{Jinwoo Kim}[orcid=0000-0003-1303-8668]
\ead{jinwookim@cbnu.ac.kr}
\cormark[1]

\cortext[1]{Corresponding author}

\begin{abstract}
Large language models (LLMs) are increasingly used in cyber threat intelligence (CTI) tasks, such as summarization and question answering. However, they may produce hallucinations, false or misleading content not grounded in the source text, which can mislead analysts and corrupt downstream security decisions. This paper presents an exploratory measurement study of hallucinations in LLM-based CTI. We introduce \ourtool{}, a reproducible framework that generates LLM outputs, extracts domain-specific entities using a fine-tuned named entity recognition model, and evaluates factual consistency using multiple similarity metrics. Using 4,940 real-world cybersecurity articles, we measure hallucinations in outputs from two commercial LLMs and one open-weight LLM. Measurements are validated against 100 human-annotated outputs and two established faithfulness detectors. Results show that semantic similarity aligns with human judgments, whereas entity-based metrics exhibit limited agreement. Differences between models are concentrated in question answering, while entity extraction quality limits what entity-level metrics can reveal.
\end{abstract}


\maketitle

\section{Introduction}
In recent years, natural language processing (NLP) has advanced considerably, largely driven by the development of large language models (LLMs). Transformer-based models, in particular, have demonstrated remarkable capabilities, enabling LLMs to generate human-like text and understand language with high accuracy~\citep{vaswani2017attention}. These advantages have facilitated the use of LLMs in tasks such as summarization, question answering, and dialogue. Prominent LLM-powered applications, including OpenAI's ChatGPT and Google Gemini, demonstrate these capabilities and have contributed to the widespread adoption of LLMs. However, these advances raise critical questions about the reliability of LLM outputs in security-sensitive operational contexts.

LLMs can also generate inaccurate information, a phenomenon known as \emph{hallucination}~\citep{ji2023survey}. Hallucination poses a significant challenge for intelligent systems because factually incorrect outputs can undermine confidence in automated workflows. This issue is particularly critical in safety- and security-critical domains such as cybersecurity, healthcare, and finance, where inaccurate outputs can mislead analysts and compromise decision-making. For instance, using LLMs for \emph{cyber threat intelligence (CTI)} has proven effective for gathering security-related data~\citep{singla2023empirical}\footnote{See also practitioner reports such as \url{https://www.linkedin.com/pulse/automating-tasks-osint-analysts-chatbot-chatgpt-deep-dive-groeneveld/} and \url{https://medium.com/@anthony.mensier/gpt-4-for-defense-specific-named-entity-extraction-47895b7fed6d}.}. Consider a scenario in which a CTI analyst collects security documents and uses an LLM-based expert system to summarize them and extract key details such as indicators of compromise (IoCs)~\citep{liao2016acing} and tactics, techniques, and procedures (TTPs)~\citep{husari2017ttpdrill}. If the LLM provides inaccurate security information, the CTI analyst may adopt misguided countermeasures. This illustrates how hallucinations can distort decision-making in cybersecurity operations. 
Thus, systematically analyzing and evaluating hallucinations in LLM-based CTI systems is essential to maintaining the integrity of security operations.

\begin{figure*}[!t]
    \centering
    \subfloat[Summary hallucination.]{
    \includegraphics[width=.5\textwidth]{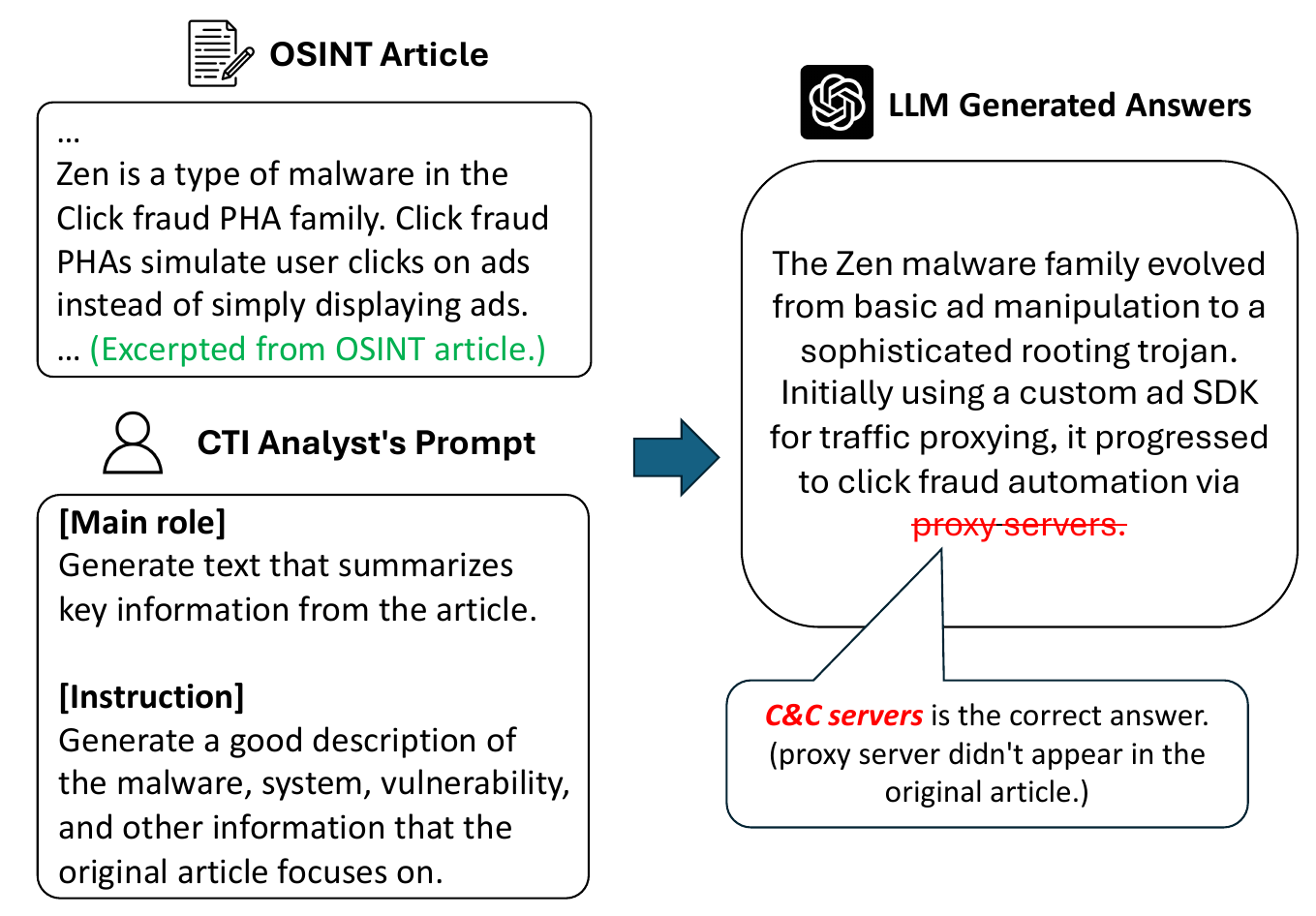}
    \label{fig:motiv_1}
    }
    \subfloat[Question \& answering (Q\&A) hallucination.]{
    \includegraphics[width=.5\textwidth]{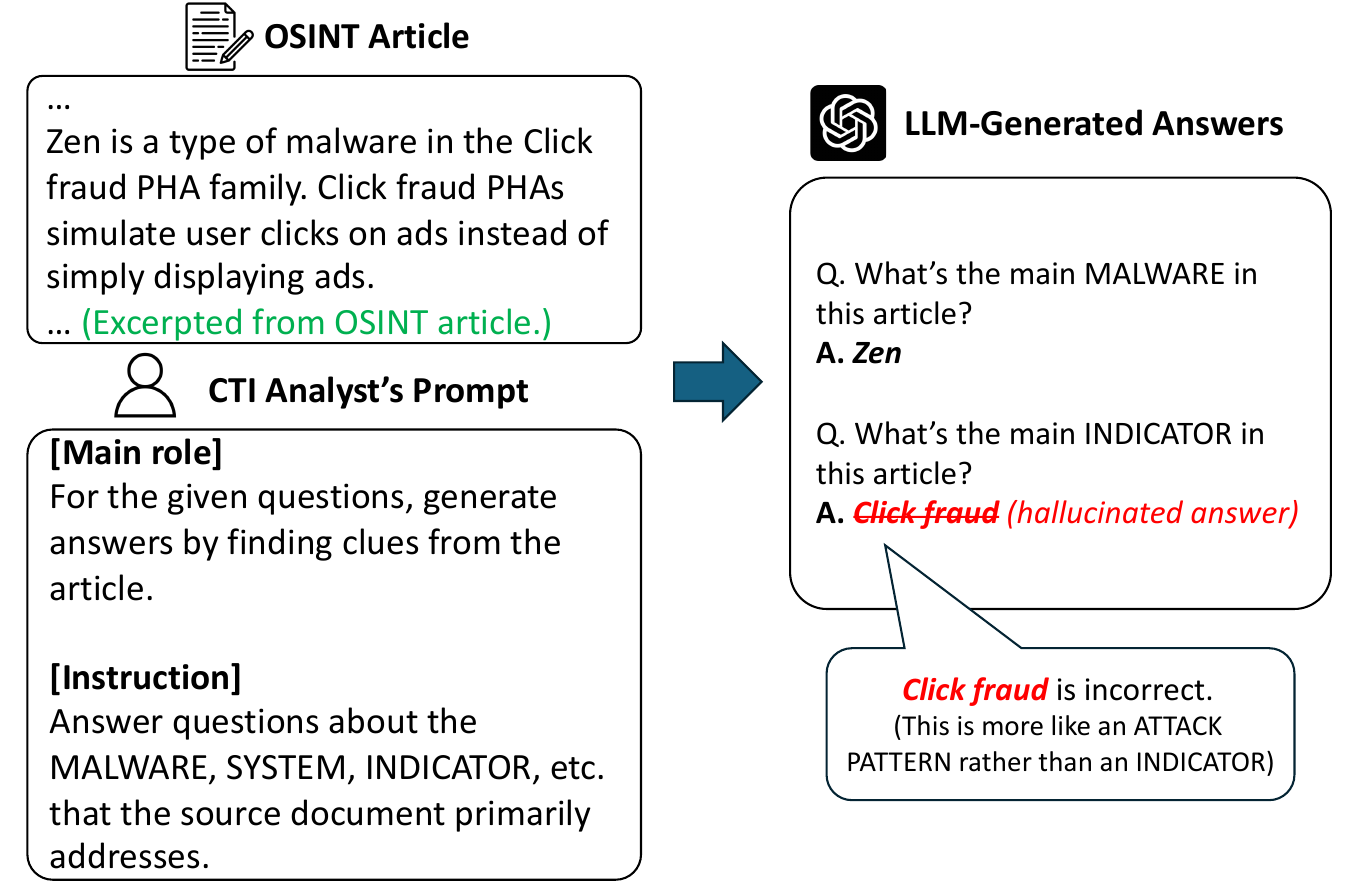}
    \label{fig:motiv_2}
    }
    \caption{Motivating examples of hallucinations in the CTI domain. In both cases, the LLM output contradicts the source article: in (a), the LLM refers to \emph{proxy servers} whereas the source article refers to \emph{command-and-control servers}; in (b), the LLM assigns \emph{click fraud} to the \texttt{Indicator} label. Both cases therefore constitute intrinsic hallucinations as defined in Section~\ref{subsec:problem}.}
    \label{fig:motiv_example}
\end{figure*}

Figure~\ref{fig:motiv_example} illustrates how hallucinations can arise when a CTI analyst relies on an LLM for \emph{summarization} and \emph{question answering} (Q\&A) to support CTI tasks. In Figure~\ref{fig:motiv_1}, the CTI analyst prompts the LLM to summarize an article about Zen malware\footnote{\url{https://security.googleblog.com/2019/01/pha-family-highlights-zen-and-its.html}}. Although most of the summary is accurate, the LLM contradicts the original article by attributing the activity to \emph{proxy servers}, whereas the source refers to \emph{command-and-control (C\&C) servers}. This constitutes an intrinsic hallucination as defined in \citep{ji2023survey}. Similarly, in Figure~\ref{fig:motiv_2}, when asked to interpret the article and answer predefined questions, the LLM correctly identifies \emph{Zen} as the \texttt{Malware} but incorrectly lists \emph{click fraud} as an \texttt{Indicator}, a label for patterns used to detect malicious activity.

In this study, we aim to conduct an exploratory measurement analysis of hallucination in LLM-based intelligent systems, focusing on their reliability in practical applications. We illustrate this through a case study in cybersecurity, where LLMs are increasingly used for CTI tasks. Our analysis focuses on summarization and Q\&A tasks, two prevalent forms of text comprehension in which hallucinations are commonly observed~\citep{ji2023survey}. Unlike prior studies that examine hallucination in general-purpose datasets~\citep{yu2023automatic,li2023halueval} or assess CTI credibility through LLM-driven claim verification~\citep{wu2026kgv}, our work is among the first to systematically measure and characterize hallucination patterns in LLM-generated CTI outputs using lightweight, reproducible metrics. Although recent work~\cite{SOOD2025110307} has proposed qualitative taxonomies of hallucination types in AI-driven cybersecurity systems, systematic large-scale empirical measurement of hallucination prevalence in real-world CTI workflows remains lacking. A concurrent study by Mezzi \emph{et al.}~\citep{mezzi2025unreliable} reports that LLMs are inconsistent and overconfident when processing full-length CTI reports, further motivating the direct comparison of their outputs with the source text.

However, conducting this measurement poses several challenges from a CTI analyst's perspective. First, collecting CTI articles, generating outputs using LLMs (e.g., summaries and answers to questions), and comparing them with the original text to evaluate similarity require substantial manual effort and are time-consuming. Second, analysts must extract malware-related entities (e.g., malware names and infection methods) from both the original articles and LLM-generated outputs, a task rendered difficult by the limited domain-specific knowledge embedded in current language models~\citep{jo2020gapfinder,jo2022vulcan}. Third, existing hallucination metrics in NLP are often unintuitive for practitioners, making it difficult to interpret and quantify the degree of hallucination.

To address these challenges, we develop a measurement framework, \ourtool{}, that collects original CTI articles and interacts with LLMs to generate corresponding outputs for hallucination analysis. \ourtool{} applies a preprocessing pipeline to clean the articles and employs a cybersecurity-specific named entity recognition (NER) model to extract malware-related entities and assign them appropriate labels. To improve entity extraction accuracy, we use a customized NER model fine-tuned for cybersecurity. Finally, \ourtool{} computes lightweight and interpretable similarity metrics comparing the original articles with the LLM-generated summaries, enabling automated and systematic detection of hallucinations based on these scores. Although existing methods such as FActScore~\citep{min2023factscore} and QAGS (Question Answering and Generation for Summarization)~\citep{wang2020asking} can evaluate factual consistency, these judge-based detectors rely on auxiliary models to evaluate each output, rendering their application computationally expensive for the large and continuously growing volume of CTI articles.

For our empirical measurement study, we curated a dataset of 4,940 security articles from Threatpost\footnote{\url{https://threatpost.com/}} and 500 articles from The Hacker News\footnote{\url{https://thehackernews.com/}}, each containing malware-related details drawn from diverse threat intelligence reports. We then used two commercial LLMs (GPT-3.5-turbo and GPT-4o-mini) and an open-weight LLM (Llama-3-8B-Instruct) to generate summaries and answers to predefined questions based on the articles. To identify ground-truth labels for malware-related entities, we fine-tuned CyNER~\citep{alam2022cyner}, a cybersecurity-specific NER model. Next, we evaluated hallucinations by applying one semantic metric (cosine similarity) and two entity-based metrics (answer matching similarity and frequency-based entity similarity). 

We present a quantitative analysis of the similarity distributions between the original articles and the LLM-generated outputs, followed by qualitative case studies of real-world hallucination examples. We further validate the measurements through a blinded human study in which 10 annotators majoring in cybersecurity or artificial intelligence rated 100 LLM outputs, a manual audit of the entity extractor, and a comparison with two established hallucination detectors, FActScore~\citep{min2023factscore} and QAGS~\citep{wang2020asking}.

Our findings are summarized as follows. First, semantic similarity aligns with human judgment, whereas the two entity-based metrics show limited agreement; therefore, our metrics are better suited for preliminary screening than for use as calibrated hallucination detectors. Second, existing judge-based detectors rarely detect hallucinations in this setting because FActScore and QAGS rate nearly every summary as faithful when the summary is generated from the CTI article used for evaluation. Third, differences between models are concentrated in the Q\&A task rather than in the summarization task. Finally, entity-level measurements depend heavily on the accuracy of the entity extractor. We quantify the extent to which entity extraction errors contribute to the observed measurement gap.

\noindent\textbf{Contributions.} Our contributions are summarized as follows:
\begin{itemize}
    \item We present an exploratory measurement study of hallucinations in LLMs applied to CTI summarization and Q\&A tasks, examining their implications for the reliability of LLM-assisted CTI workflows.
    \item We develop \ourtool{}, a framework that automates the hallucination analysis pipeline, including article collection, LLM interaction, entity extraction, and similarity-based evaluation.
    \item We conduct extensive experiments on a curated dataset of 4,940 security articles, using quantitative analyses and real-world case studies to quantify the frequency of hallucinations, the entity types affected, and the tasks in which they occur, while highlighting their potential impact on CTI operations.
    \item To promote reproducibility, we publicly release our code\footnote{\url{https://github.com/SeungHo0422/HalluVision-official}}, dataset\footnote{\url{https://doi.org/10.17632/z8sfdnch2c}}, and the fine-tuned CyNER checkpoint\footnote{\url{https://huggingface.co/Ricky-dev/halluvision-cyner-xlmr-base}}, enabling other researchers and practitioners to replicate our findings and build upon the framework.
\end{itemize}

The remainder of this paper is organized as follows. Section~\ref{sec:background} reviews the background of CTI and hallucination and presents the problem statement, while Section~\ref{sec:design} describes \ourtool{}’s  measurement methodology. Section~\ref{sec:eval} presents the measurement results and validation studies. Sections~\ref{sec:discussion} and~\ref{sec:limitations} discuss the findings and their limitations, respectively, while Sections~\ref{sec:related} and~\ref{sec:conclusion} review related work and present the conclusions, respectively.

\section{Background and Motivation}
\label{sec:background}

This section provides background information and explains the motivation for  our study.

\subsection{Cyber Threat Intelligence}

CTI involves a systematic approach to collecting and analyzing large amounts of cyber threat data to identify sophisticated cyberattacks and develop effective countermeasures~\citep{tounsi2018212}. For example, this approach involves collecting information about malware that exploits zero-day vulnerabilities, including signatures and IP addresses, and providing CTI analysts with IoCs for defensive purposes~\citep{liao2016acing}. Within the realm of CTI, open source intelligence (OSINT) is a specialized field that focuses on collecting information from publicly available sources such as social media, blogs, and articles. LLM applications such as ChatGPT are increasingly being used in OSINT~\citep{singla2023empirical}\footnote{See also practitioner reports such as \url{https://blog.sociallinks.io/using-the-power-of-chatgpt-for-osint/} and \url{https://www.linkedin.com/pulse/automating-tasks-osint-analysts-chatbot-chatgpt-deep-dive-groeneveld/}.}. This trend is attributed to the greater speed and precision of LLM-based text processing compared with those of conventional human methods. Tasks such as text summarization and question answering, at which LLMs excel, mirror the tasks humans perform when collecting malware information through OSINT.

\subsection{Information Gap and Hallucination}

CTI also faces a challenge known as the \emph{information gap}, characterized by conflicting data from different knowledge sources. For example, conflicting reports identify \emph{February 2011} and \emph{September 2011} as the discovery date of the Shylock malware~\citep{jo2020gapfinder}. Here, the information gap refers to discrepancies between external CTI sources. Language models, including LLMs, can also exhibit an information gap termed \emph{hallucination}, in which they generate inaccurate information that deviates from established knowledge sources~\citep{ji2023survey}. In cybersecurity, this means that LLMs may distort malware information and present it differently from the original sources. LLM hallucinations are particularly difficult to detect in practice. Although the source article is available, manually comparing every generated output with the source is not scalable because it requires the same effort analysts delegate to LLMs. This motivates the automated measurement proposed in this study. Thus, although both phenomena can reduce the reliability of CTI, our study focuses exclusively on hallucinations in LLM outputs.

\subsection{Problem Statement}
\label{subsec:problem}

A \emph{security article} is defined as unstructured text that provides details about malware, such as its name, infection method, and reported date. Although these articles may cover diverse topics, this paper focuses specifically on \emph{malware}, assuming that CTI analysts seek to stay informed about emerging threats to devise effective countermeasures. A major challenge is that security articles often contain lengthy, unstructured text, making the extraction of relevant threat vectors time-consuming. To streamline this process, CTI analysts require concise and accurate information from OSINT sources, with support from LLMs.

\begin{figure*}[!t]
    \centering
    \includegraphics[width=.95\textwidth]{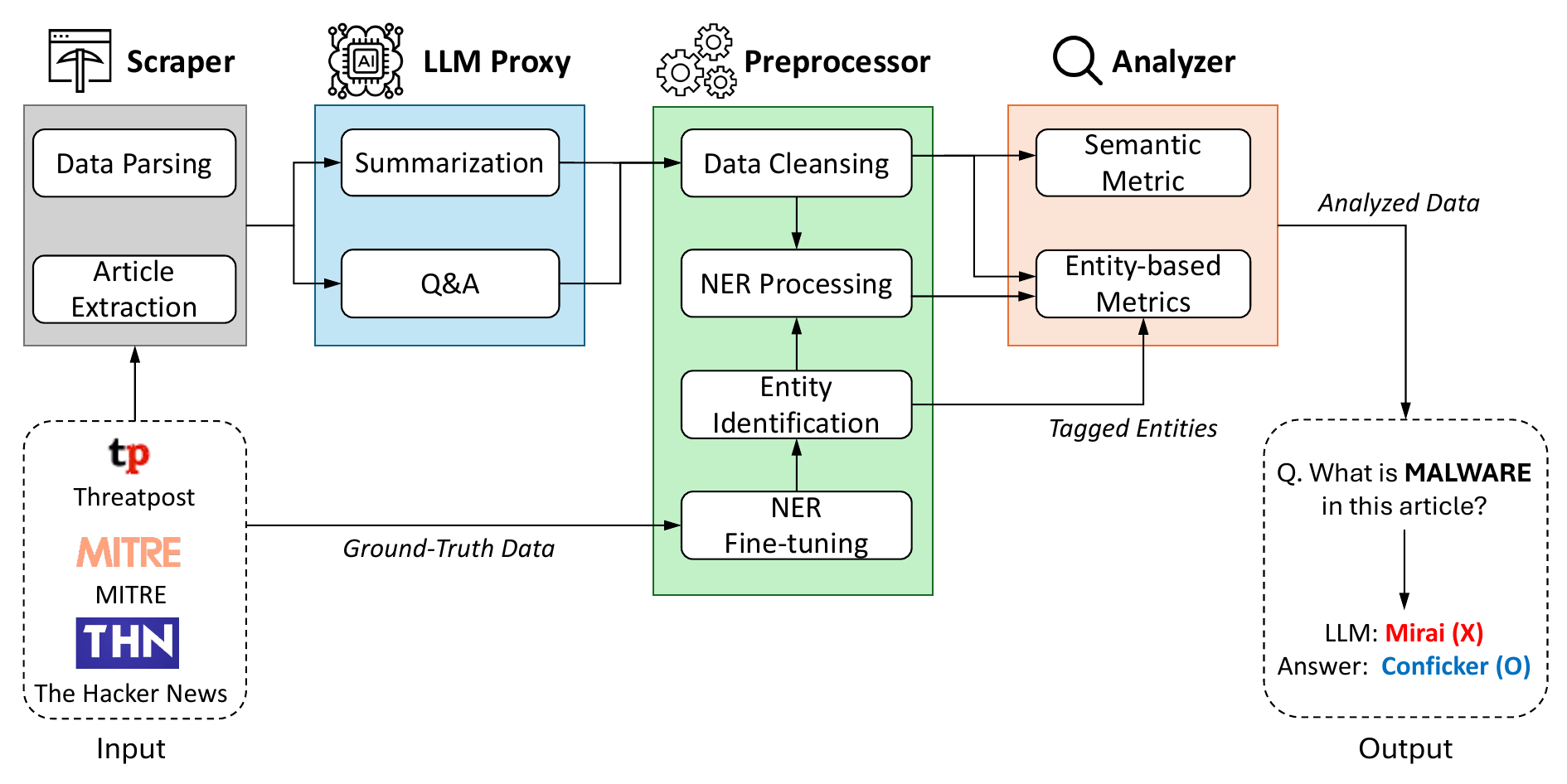}
    \caption{Our measurement framework, \ourtool{}, comprising four components: Scraper, LLM Proxy, Preprocessor, and Analyzer.}
    \label{fig:overview}
\end{figure*}

A \emph{summary} is a condensed version of a text, typically one or two sentences, that captures essential information to aid practitioners in defense planning. LLMs can generate such summaries, referred to as \emph{LLM-generated summaries}, using prompts such as “\emph{Summarize the given malware article concisely.}” LLMs can also answer predefined questions (e.g., “\emph{When was the malware discovered?}” or “\emph{Which vulnerability does the malware exploit?}”) through their \emph{Q\&A} feature, providing CTI analysts with critical information. We refer to these responses as \emph{LLM-generated answers}.

In this context, we propose that an information gap may arise between a security article and its LLM-generated summary. Hallucinations within the summary can introduce incorrect details about malware, undermining the CTI analyst's objectives. We refer to this phenomenon as \emph{summary hallucination}. Similarly, inaccuracies in LLM-generated responses to predefined questions are termed \emph{Q\&A hallucination}. Although these issues are well-documented in NLP~\citep{ji2023survey}, our focus is on assessing their relevance and impact in cybersecurity.

\section{Measurement Methodology}
\label{sec:design}

In this section, we provide an overview of our measurement framework and describe the methods used to assess hallucinations in the cyber threat intelligence domain.

\subsection{HalluVision Overview}

Identification of summary and Q\&A hallucinations requires evaluating information consistency between texts obtained from knowledge sources (i.e., OSINT) and those generated by LLMs (i.e., LLM-generated summaries and LLM-generated answers). To this end, we present \ourtool{}, a systematic framework comprising four key components: \emph{scraper}, \emph{LLM proxy}, \emph{preprocessor}, and \emph{analyzer}, as shown in Figure~\ref{fig:overview}.

\noindent\textbf{Scraper.} The scraper collects security-related articles from OSINT sources. Modern websites often rely on JavaScript-based dynamic rendering, which poses challenges for static web crawlers that cannot process dynamically generated content. Dynamic web crawlers are therefore needed to collect data from these websites. We use the Selenium library to scrape dynamically generated pages. This approach enables us to access individual articles and extract their titles and body text by parsing the HTML. By focusing on these elements, we collect only the information relevant to the analysis. The extracted texts are then stored in a database for further processing.

\noindent\textbf{LLM Proxy.} The LLM proxy interacts with a target LLM by sending prompts containing the original text and receiving the corresponding responses. It performs two key tasks: \emph{Summarization} and \emph{Question \& Answering} (Q\&A). The summarization task generates concise summaries by prompting the LLM to condense the original text while highlighting its essential points. The Q\&A task focuses on extracting specific information from the original text using predefined Q\&A prompts. These questions are designed to clarify the main points covered in the text. The LLM-generated outputs are stored in a database for subsequent analysis and comparison with the original texts.

\noindent\textbf{Preprocessor.} The preprocessor then refines the data for the analyzer. This step is particularly important for abstractive summarization, which, unlike Q\&A, requires extracting meaningful entities from both the original content and its summaries using NER techniques. NER involves identifying and classifying entities in text according to predefined labels~\citep{10.1109/tkde.2020.2981314}. For general text, labels such as \texttt{Person}, \texttt{Organization}, \texttt{Location}, and \texttt{Date} can be used, whereas domain-specific text requires appropriate labels. CTI-related texts contain specialized terminology, requiring customized labels. The OASIS Open Foundation categorizes information relevant to CTI into Structured Threat Information Expression (STIX) objects\footnote{\url{https://oasis-open.github.io/cti-documentation/stix/intro.html}}. In our study, we use this STIX categorization to identify NER labels for extracting entities from OSINT sources. During NER, the preprocessor classifies entities according to predefined labels and refines the extracted information. This approach helps capture the semantic content of the text through effective entity identification and extraction. The NER model is a cybersecurity-specific model (CyNER) fine-tuned on human-annotated CTI articles (Section~\ref{subsec:ex_setup}).

\noindent\textbf{Analyzer.} Finally, the analyzer compares the LLM outputs with their source articles. Summaries are assessed with one semantic metric (cosine similarity between text embeddings) and two entity-based metrics (answer matching similarity and frequency-based entity similarity), whereas Q\&A answers are assessed with answer matching similarity. Low scores flag outputs that may contain hallucinations. The following subsections describe \ourtool{} in detail.

\subsection{LLM Text Generation}

\begin{table*}
    \footnotesize
    \centering
    \caption{Structure of the prompts used for summarization and question answering. The complete prompts are provided in Appendix~\ref{appendix:prompts}.}
    \begin{tabular}{p{2cm} p{6cm} p{6cm}}
        \toprule
        \textbf{Type} & \textbf{Summarization Prompt} & \textbf{Q\&A Prompt} \\
        \midrule
        \textbf{Role} &
        NER expert in the CTI domain &
        NER expert in the CTI domain \\ \midrule
        
        \textbf{Instructions} &
        - Summarize blog content in 50–70 words. \newline
        - Use only words and content from the original text. &
        - Answer five predefined entity-based questions. \newline
        - Use short answers (ideally one entity). \newline
        - Output ``no answer'' if info is missing. \\ \midrule
        
        \textbf{Input} &
        \texttt{\{original\_text\}} &
        \texttt{\{original\_text\}} \\ \midrule
        
        \textbf{Output} &
        Single summary paragraph &
        5 question–answer pairs \\
        \bottomrule
    \end{tabular}
    \label{tab:prompts}
\end{table*}

The extracted text from the source articles is sent to the target LLM along with customized prompt instructions. To facilitate this, we designed an LLM proxy that generates prompts by embedding a query tailored to the specific task (e.g., summarization or Q\&A) alongside the source text. To optimize responses for capturing essential knowledge, we apply task-specific prompt engineering. These prompts are used by two specialized modules in the LLM proxy: the summarizer and the inquirer. Table~\ref{tab:prompts} provides the prompt structures. We deliberately keep the role persona identical across the two tasks, so that any difference between summarization and Q\&A is attributable to the task rather than the role instruction.

\noindent\textbf{Summarization.} The LLM proxy’s summarization task generates summaries containing the key content of the original articles extracted by the scraper. To ensure a fair analysis, we design our summarization prompt to satisfy the following conditions: First, the summary text should contain only words from the original text, excluding additional responses or opinions generated by the model.
Second, the prompt limits the maximum number of words to ensure an appropriate summary length.
These LLM-generated summaries are also stored in a database for subsequent comparison with the original article.

\noindent\textbf{Question \& Answering.} 
For Q\&A, the LLM proxy generates a prompt containing a series of questions alongside a given article and sends it to a target LLM to retrieve answers. The prompt is carefully designed to account for the following factors:

First, we identify the key labels commonly referenced in malware-related articles. This step ensures consistency when comparing NER entities extracted from the generated answers with those in the original sources. Specifically, we define five labels: \texttt{Malware}, \texttt{System}, \texttt{Indicator}, \texttt{Vulnerability}, and \texttt{Organization}. These labels align with those specified in CyNER~\citep{alam2022cyner}, which serves as our baseline model. Based on an extensive analysis of malware-related articles, these five labels are validated to effectively capture the most critical information.

\begin{table*}[t]
    \centering
    \caption{Examples of abstractive summarization and question \& answering outputs generated by an LLM from an article.}
    \footnotesize
    \resizebox{\textwidth}{!}{
    \begin{tabular}{p{4cm} p{13cm}}
        \toprule
        \textbf{Source} & \textbf{Content} \\
        \midrule
        \textbf{Original Article} & 
        300 restaurants and at least 50,000 payment cards compromised by two separate campaigns against MenuDrive, Harbortouch and InTouchPOS services. \textcolor{blue}{\textbf{Magecart}} campaigns have been skimming payment-card credentials of unsuspecting customers using three online restaurant-ordering systems, affecting about 300 restaurants that use the services and compromising tens of thousands of cards so far, researchers have found. Two separate ongoing Magecart campaigns have injected e-skimmer scripts into the online ordering portals of restaurants using three separate platforms: \textcolor{LimeGreen}{\textbf{MenuDrive}}, Harbortouch, and InTouchPOS, researchers from \textcolor{Orange}{\textbf{Recorded Future}} revealed in a blog post this week.
        \vspace{0.1in}
        (...)
        \vspace{0.1in}
        ``This campaign likely began no later than Jan. 18, 2022, and as of this report, a portion of the restaurants remained infected,'' they noted in the post. However, the malicious domain used for the campaign, which researchers identified as \textcolor{red}{\textbf{authorizen[.]net}}, has been blocked since May 26, they said. \\
        \midrule
        \textbf{Abstractive Summarization} & 
        Two \textcolor{blue}{\textbf{Magecart}} campaigns have compromised over 50,000 payment cards from approximately 300 restaurants using \textcolor{LimeGreen}{\textbf{MenuDrive}}, Harbortouch, and InTouchPOS services. Researchers from Recorded Future reported that e-skimmer scripts were injected into these platforms, exposing customers' payment card data and personal information. The campaigns began in late 2021 and early 2022, with ongoing infections and malicious domains still active. Smaller restaurant platforms remain attractive targets for cybercriminals due to their vulnerabilities. \\
        \midrule
        \textbf{Question \& Answering} & 
        Q1. What is the main Malware?: \textcolor{blue}{\textbf{Magecart}} \newline
        Q2. What is the main System?: \textcolor{LimeGreen}{\textbf{MenuDrive}} \newline
        Q3. What is the main Indicator?: \textcolor{red}{\textbf{authorizen[.]net}} \newline
        Q4. What is the main Vulnerability?: no answer \newline
        Q5. What is the main Organization?: \textcolor{Orange}{\textbf{Recorded Future}} \\
        \bottomrule
    \end{tabular}
    }
    \label{tab:hallucination_text_comparison}
\end{table*}

Second, we design prompts to guide the LLM in accurately answering questions related to predefined labels. These prompts include detailed instructions to ensure concise and precise responses. The LLM is instructed to provide only the entity itself, rather than a full sentence, in response to clearly phrased questions about whether an entity corresponding to a specific label is present in the article. If no information related to a label is found in the article, the LLM is instructed to respond with ``no answer'' to prevent it from generating incorrect inferences\footnote{Following the guidance in \url{https://github.com/dair-ai/Prompt-Engineering-Guide}.}. Table~\ref{tab:hallucination_text_comparison} shows the summary and answers generated for one article.

\subsection{Preprocessing}

The preprocessor standardizes text formats through data cleansing and identifies named entities to facilitate hallucination analysis in subsequent stages.

\noindent\textbf{Data Cleansing.}
Articles and responses obtained from the scraper and LLM proxy must be converted into a structured format for analysis. This involves preprocessing steps such as converting text to lowercase, removing unnecessary terms and punctuation, extracting root words, and applying lemmatization. Irrelevant phrases, including promotional text such as \texttt{"Share this article:"}, are removed during data cleansing. After cleansing, an NER model extracts meaningful entities to identify security-related entities from both the original and generated texts. The LLM proxy operates in two modes: summarization and Q\&A. As the Q\&A mode inherently includes named entities in its output, no additional NER processing is required. Consequently, preprocessing focuses on LLM-generated summaries, for which NER is used to extract entities.

\noindent\textbf{Entity Identification.}
Hallucinations in CTI-related OSINT articles pose a significant risk, particularly when they involve information directly linked to cyber threats, such as malware and vulnerabilities. Therefore, our analysis prioritizes hallucinations in these areas over general topics. To achieve this, we define five entity labels and extract corresponding entities from both source and LLM-generated texts using NER. A hallucination is detected when entities in the LLM-generated text do not align with those in the source text. The label definitions are based on descriptions from STIX and CyNER~\citep{alam2022cyner}. Table~\ref{tab:label_definition} summarizes these definitions.

\begin{table*}[t]
\centering
\footnotesize
\caption{Entity categories considered in our study.}
\label{tab:label_definition}
\begin{tabular}{p{2.5cm} p{10cm}}
\toprule
\textbf{Entity Type} & \textbf{Description} \\
\midrule
Malware & The name of the primary malware discussed in the article. If no specific name is provided, it may refer to a general malware category. \\ \midrule
System & Operating systems (e.g., Linux, Windows), software (e.g., Chrome, Slack), or hardware leveraged by threat actors in attacks. \\ \midrule
Indicator & Patterns used to detect malicious activity, such as domain names, URLs, IP addresses, filenames, hashes, email addresses, and port numbers. \\ \midrule
Vulnerability & Common Vulnerabilities and Exposures (CVE) identifiers (e.g., CVE-2024-9264) and exploit mentions (e.g., command injection vulnerabilities) explicitly stated in the article. \\ \midrule
Organization & Entities directly affected by cyber threats discussed in the article. Organizations that merely observed or reported the threats are excluded. \\
\bottomrule
\end{tabular}
\end{table*}

\subsection{Hallucination Analysis}
\label{subsec:hallu_anal}

The analyzer evaluates the degree of hallucination in LLM-generated texts by comparing them with the original source text. LLM hallucinations are broadly categorized into two types: (i) \emph{intrinsic} and (ii) \emph{extrinsic}~\citep{ji2023survey}. Intrinsic hallucinations occur when LLM-generated content contradicts factual information (e.g., Figure~\ref{fig:motiv_example}), whereas extrinsic hallucinations arise when the model produces content that is not directly supported by factual data. Our work primarily focuses on intrinsic hallucinations, because extrinsic hallucinations cannot be directly verified. Intrinsic hallucinations can be particularly problematic in cybersecurity applications; for instance, if an LLM incorrectly describes malware characteristics, flawed security measures could be implemented. 

\begin{table*}[t]
    \footnotesize
    \centering
    \caption{Hallucination metrics used in our study and their interpretations. 
    AS = Abstractive Summarization, Q\&A = Question \& Answering, NER = Named Entity Recognition.}
    \resizebox{\textwidth}{!}{
    \begin{tabular}{p{3.5cm} c c c c}
        \toprule
        \textbf{Metric} & \textbf{Preprocessing Required} & \textbf{Applicable To} & \textbf{Interpretation of High Score} & \textbf{Interpretation of Low Score} \\ 
        \midrule

        Cosine Similarity & Sentence Embedding & AS & Semantically consistent with the source & Semantic deviation or abstraction \\ \midrule

        Answer Matching Similarity – Precision & NER & AS, Q\&A & Relevant entities correctly included & Irrelevant or spurious entities included \\ \midrule

        Answer Matching Similarity – Recall & NER & AS, Q\&A & Most reference entities are retrieved & Key entities are missing \\ \midrule
        
        Frequency-based Entity Similarity & NER & AS & Similar entity-type distribution & Topic drift or entity imbalance \\ 
        \bottomrule
    \end{tabular}
    }
    \label{tab:hallucination_metrics_full}
\end{table*}

Formally, we aim to determine whether the LLM-generated text $Y$ aligns with the original text $X$. To assess this alignment, we employ three similarity metrics: (i)~cosine similarity, (ii)~answer matching similarity, and (iii)~frequency-based entity similarity. The definitions and interpretations are summarized in Table~\ref{tab:hallucination_metrics_full}. Although our evaluation metrics are relatively straightforward, we intentionally selected them for their interpretability and robustness in the domain-specific CTI context.
Compared with complex state-of-the-art hallucination scorers, these metrics provide greater transparency and consistency for practitioners evaluating LLM outputs in high-stakes applications. Note that worked examples of the three metrics are provided in Appendix~\ref{appendix:examples} and a rule that combines the three metrics into a single score is examined in Appendix~\ref{appendix:combined}.

\noindent\textbf{Cosine Similarity.}
Cosine similarity measures the similarity between two vectors by computing the cosine of the angle between them. To calculate this metric, we convert both the original and LLM-generated texts into embedding vectors and compute their cosine similarity. For this evaluation, we use \emph{jina-embeddings-v3}~\citep{sturua2024jinaembeddingsv3multilingualembeddingstask}, a top-performing state-of-the-art model for semantic textual similarity (STS) tasks on the Hugging Face Massive Text Embedding Benchmark (MTEB) leaderboard~\citep{muennighoff2023mtebmassivetextembedding}. The cosine similarity $C$ is defined as
\begin{equation}
    C(X,Y)=\cos(\theta)=\frac{X' \cdot Y'}{\|X'\|\|Y'\|},
\end{equation}
where $X'$ and $Y'$ are the embedding vectors generated by the embedding model from texts $X$ and $Y$, respectively.

\noindent\textbf{Answer Matching Similarity.}
This metric evaluates hallucinations by comparing key entities associated with predefined labels in the source text and the LLM-generated text. This approach aims to assess whether the information extracted from the LLM output matches that in the original text. A prediction is considered correct if the predicted entity appears under the same label in the source text. Entities extracted from the original text are thus checked for consistency with those found in the generated output. We compute precision and recall to quantify this similarity. Precision is defined as
\begin{equation}
\text{Precision} = \frac{TP}{TP + FP},
\end{equation}
where $TP$ (true positive) denotes the number of entities from the original text that are correctly included in the LLM-generated text, and $FP$ (false positive) denotes the number of entities included in the LLM-generated text that do not appear in the original text. Lower precision indicates that the LLM tends to include information not grounded in the source, consistent with the definition of hallucination.
Recall is defined as
\begin{equation}
\text{Recall} = \frac{TP}{TP + FN},
\end{equation}
where $FN$ (false negative) denotes the number of entities from the original text that are not included in the LLM-generated text. Lower recall indicates that the LLM fails to capture or reproduce information present in the original text.

\noindent\textbf{Frequency-based Entity Similarity.}
This metric compares the distribution of named entities across predefined labels in the original text and the LLM-generated output.  
For each document, we count the number of entities corresponding to each predefined label $i$ in both the original article and the LLM-generated text, denoted by $c_i^X$ and $c_i^Y$, respectively.  
We then normalize these counts to form probability distributions:
\begin{equation}
p_i^X = \frac{c_i^X}{\sum_{j=1}^{n} c_j^X}, \quad
p_i^Y = \frac{c_i^Y}{\sum_{j=1}^{n} c_j^Y}
\end{equation}
where $n$ is the number of predefined entity labels.

To quantify the alignment between the two distributions, we compute the frequency-based entity similarity $F$ as follows:
\begin{equation}
\text{F}(X, Y) = 1 - \frac{1}{2} \sum_{i=1}^{n} \left| p_i^X - p_i^Y \right|
\end{equation}
The term $\sum_{i=1}^{n} \left| p_i^X - p_i^Y \right|$ corresponds to the $L_1$ distance between the two distributions.  
We divide this value by 2 to ensure that the similarity score is bounded between 0 and 1, with 1 indicating perfect alignment and 0 denoting complete divergence.

\section{Results}
\label{sec:eval}

In this section, we present the results of our study, addressing the following research questions (RQs):

\begin{itemize}
    \item \textbf{RQ1.} How closely do LLM-generated outputs resemble the original
    articles, as measured by our three proposed similarity metrics?
    (Section~\ref{subsec:quantitave})

    \item \textbf{RQ2.} Do the measurements hold beyond a single model and a
    single source? (Sections~\ref{appendix:open-weight} and~\ref{subsec:second-source})

    \item \textbf{RQ3.} Do the measurements agree with human judgment and
    existing hallucination detectors?
    (Sections~\ref{subsec:human_validation} and \ref{subsec:baselines})

    \item \textbf{RQ4.} How much of the measured error is attributable
    to the NER ground truth rather than the LLM?
    (Sections~\ref{subsec:ner_reliability} and ~\ref{subsec:decomp})

    \item \textbf{RQ5.} Do our measurements reveal concrete instances of hallucination
    in LLM-generated outputs? (Section~\ref{subsec:case_studies})
\end{itemize}

\subsection{Dataset}
To compile a diverse dataset of CTI articles and blog posts, we developed a dynamic web crawler in Python using Selenium. This approach enabled efficient extraction of article titles and main content from the HTML, supporting both static and dynamically rendered webpages. The extracted text was stored locally and subsequently processed through a preprocessing pipeline before being used as input to the language models evaluated in this study.

Specifically, we collected 4,940 CTI articles from Threatpost for evaluation, along with 50 additional articles from the MITRE ATT\&CK website\footnote{\url{https://attack.mitre.org/}} to fine-tune our ground-truth model. Table~\ref{tab:scrapdata} summarizes the dataset. We selected these articles based on the malware families listed as examples by MITRE to ensure a balanced distribution of our five target labels. The ground-truth data for these articles were generated through a human annotation survey in which participants were instructed to identify and list all corresponding entities. We also collected 500 articles from The Hacker News, using stratified sampling based on its article labels and a fixed seed. This set is used only for the transferability check reported in Section~\ref{subsec:second-source} and was not used during framework development.

\begin{table*}
    \footnotesize
    \centering
    \caption{Summary of scraped articles and generated outputs by task, model, and corpus. Dashes indicate combinations that were not run.}
    \label{tab:scrapdata}
    \begin{tabular}{ c c c c }
        \toprule
        & & \multicolumn{2}{c}{\textbf{\# of Articles}} \\
        \cmidrule(lr){3-4}
        \textbf{Task} & \textbf{Source / Model} & \textbf{Threatpost} & \textbf{The Hacker News} \\
        \midrule
        \emph{Article Scraping}     & ---   & 4,940 & 500 \\
        \emph{NER Fine-tuning Data} & MITRE & \multicolumn{2}{c}{50} \\
        \midrule
        \multirow{3}{*}{\emph{Abstractive Summarization}}
            & gpt-4o-mini                & 4,940      & 500 \\
            & gpt-3.5-turbo              & 4,940      & 500 \\
            & Llama-3-8B-Instruct & 500 & --- \\
        \midrule
        \multirow{3}{*}{\emph{Q\&A}}
            & gpt-4o-mini                & 4,935      & 500 \\
            & gpt-3.5-turbo              & 4,936      & 500 \\
            & Llama-3-8B-Instruct & 500 & --- \\
        \bottomrule
    \end{tabular}
\end{table*}

Using the OpenAI API, we generated responses from two target LLMs: \texttt{gpt-3.5-turbo} and \texttt{gpt-4o-mini}. Of the 4,940 documents, at most five yielded invalid answers for the Q\&A task because of in-context learning issues (discussed in Section~\ref{subsec:case_studies}). Given that these failures accounted for about 0.1\% of the dataset, their impact on the overall evaluation was considered negligible, and the corresponding outputs were excluded from the reported results. We also used one open-weight model, \texttt{Meta-Llama-3-8B-Instruct}, on a subset of 500 Threatpost articles rather than the full corpus. Running an 8B model on all 4,940 articles is feasible with the same code; however, a stratified subset is sufficient to determine whether the framework and its conclusions generalize beyond a single vendor. Articles are grouped into quartiles based on the number of gold entities extracted from the source text and sampled uniformly across groups with a fixed seed. All LLM configuration parameters are presented in Appendix~\ref{appendix:api-config}.

\subsection{Experimental Setup}
\label{subsec:ex_setup}

We conducted our experiments on a machine equipped with an Intel i9-12900K CPU, 64 GB of RAM, and an NVIDIA GeForce RTX 3090 GPU. All implementations were developed in Python 3.9 using the PyTorch library.

\noindent\textbf{Ground-Truth Model.} 
Our similarity metrics primarily rely on NER to identify entities in either the original text or the LLM-generated text. 
To verify whether the generated text contains correct entities, access to gold-standard labels is essential.
However, manually annotating ground-truth labels is time-consuming and prone to human error. 
Therefore, we use CyNER~\citep{alam2022cyner}, a cybersecurity-specific NER model based on XLM-RoBERTa, as our automatic labeling tool. 
To enhance its accuracy and reliability, we fine-tuned CyNER using the AdamW optimizer for 30 epochs, with a maximum sequence length of 128 and a learning rate of 5e-6, following the training settings suggested in the original CyNER implementation. 
Our fine-tuned model achieved 74.69\% precision, 77.84\% recall, and a 76.23\% F1-score, indicating reasonably strong performance for downstream hallucination evaluation. 

\begin{figure}
    \centering
    \includegraphics[width=\linewidth]{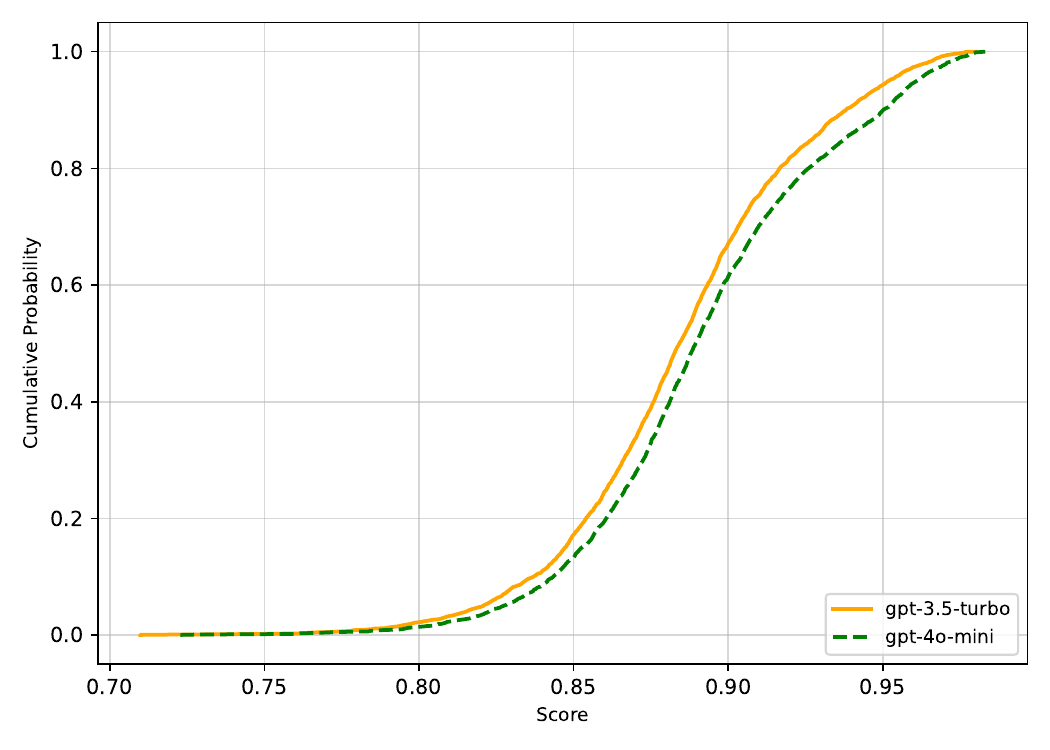}
   \caption{Cumulative distribution function (CDF) of cosine similarities between original articles and their LLM-generated summaries. Lower CDF curves (toward the bottom right) indicate better semantic similarity, as more samples achieve higher cosine similarity scores.}
    \label{fig:CosineSimilarity}
\end{figure}

\subsection{Metric Measurement}
\label{subsec:quantitave}

We first performed a quantitative analysis by examining the distributions of each similarity metric to assess the prevalence of hallucinations in LLM-generated text. 

\subsubsection{Abstractive Summarization}
We evaluated information matching between original texts and LLM-generated summaries using cosine, answer matching, and frequency-based entity similarity. Specifically, for the two entity-based metrics, we performed NER on both the source text and the summary using our ground-truth model.

\begin{figure}
    \centering
    \subfloat[Precision]{
        \includegraphics[width=\linewidth]{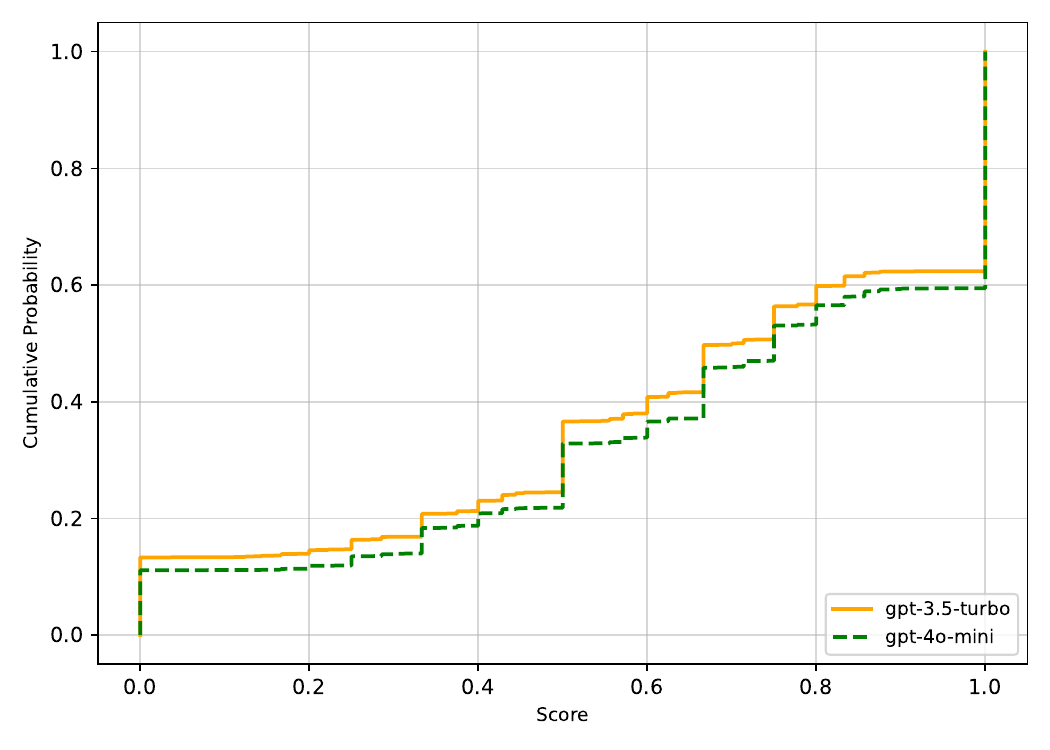}
        \label{fig:summary_prec}
    }
    \hfill
    \subfloat[Recall]{
        \includegraphics[width=\linewidth]{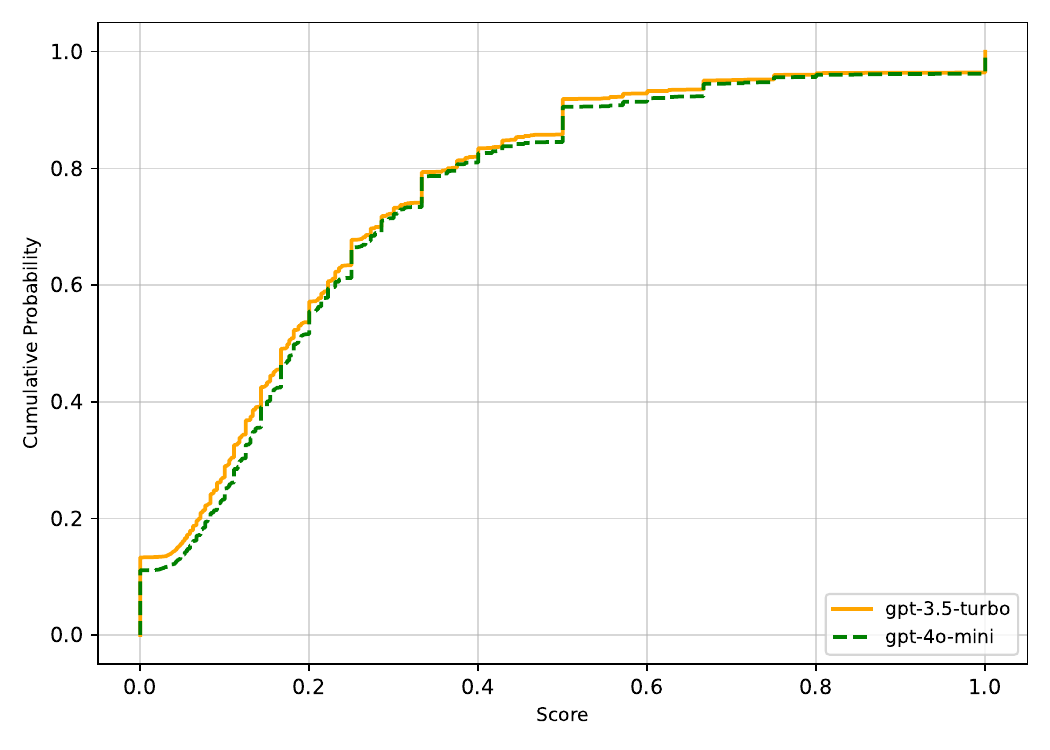}
        \label{fig:summary_recall}
    }
    \caption{CDF of answer matching similarities between original articles and their LLM-generated summaries. Lower CDF curves (toward the bottom right) indicate better similarity, as more samples achieve higher precision or recall.}

    \label{fig:prec-recall-cdf}
\end{figure}

\noindent\textbf{Cosine Similarity.}
Figure~\ref{fig:CosineSimilarity} presents the cumulative distribution function (CDF) of cosine similarity scores between 4,940 security-related documents and their summaries generated by \texttt{gpt-3.5-turbo} and \texttt{gpt-4o-mini}.
Overall, more than 40\% of the summaries exhibit high similarity scores exceeding 0.9, whereas approximately 20\% score below 0.85\footnote{Note that this cutoff does not directly indicate a hallucination threshold. Identifying thresholds remains a topic for future research (Section~\ref{sec:limitations}).}.
However, because cosine similarity measures overall semantic similarity based on text embeddings, further qualitative analysis is needed to confirm the presence of hallucinated content.
Notably, \texttt{gpt-4o-mini} consistently achieves higher similarity scores than \texttt{gpt-3.5-turbo}, aligning with expectations given its more advanced architecture.

Although high cosine similarity, computed over 1,024-dimensional dense embeddings, suggests a low degree of information loss between the source and summary, it does not guarantee factual consistency. As demonstrated in Section~\ref{subsec:case_studies}, hallucinations can still arise in localized content even when the general similarity exceeds 79.35\%. This highlights the need to combine semantic measures with more targeted metrics for robust hallucination detection.

\noindent\textbf{Answer Matching Similarity.} 
We analyzed the answer matching similarity between the LLM-generated summaries and the source texts for 4,940 articles. For \texttt{gpt-4o-mini}, the average precision is 0.702, while the average recall is relatively low at 0.1822. The high precision indicates a low number of FPs, meaning that the LLM-generated text is less likely to contain entities that do not exist in the original text. In contrast, the low recall indicates that many entities are missing from the LLM-generated summaries, leading to a high number of FNs. This high number of FNs can be attributed to the inherent nature of the summarization task. Because summaries are expected to be concise and contain only the most salient information, many entities present in the original text are naturally omitted from the generated summary. Consequently, a significant portion of the ground-truth entities are missing from the prediction, resulting in low recall.

Figure~\ref{fig:prec-recall-cdf} shows the cumulative distributions of precision and recall for \texttt{gpt-3.5-turbo} and \texttt{gpt-4o-mini}.  
Approximately 40\% of the LLM-generated summaries achieve a precision score of 1.0, while the remaining 60\% fall below 1.0 (Figure~\ref{fig:summary_prec}).  
In contrast, recall scores are notably lower: as shown in Figure~\ref{fig:summary_recall}, most summaries fail to reach a perfect recall of 1.0, suggesting that key entities from the original articles are frequently omitted.  

Overall, recall is consistently lower than precision across both models, reinforcing the observation that although the generated entities are generally accurate, important information is often missing. Despite this, the relatively high precision indicates that most entities appearing in the summaries can also be found in the corresponding source documents.  
Between the two models, \texttt{gpt-4o-mini} consistently outperforms \texttt{gpt-3.5-turbo} in both precision and recall, demonstrating better entity preservation and summary fidelity.  
This trend is consistent with the cosine similarity results, further underscoring the superior performance of \texttt{gpt-4o-mini}.

\begin{table*}[!t]
    \footnotesize
    \centering
    \caption{Label-wise answer matching similarities in abstractive summarization and question answering.}
    \label{tab:AnswerMatchingLabelwise}
    \begin{tabular}{l l l c c c c c}
        \toprule
        \textbf{Task} & \textbf{Model} & \textbf{Metric} & \textbf{Malware} & \textbf{System} & \textbf{Indicator} & \textbf{Vulnerability} & \textbf{Organization} \\
        \midrule
        \multirow{4}{*}{\emph{Abstractive Summarization}}
            & \multirow{2}{*}{gpt-4o-mini}
                & Precision & 0.7401 & 0.7187 & 0.4977 & 0.1775 & 0.7641 \\
            & & Recall      & 0.2241 & 0.1924 & 0.0315 & 0.0717 & 0.1786 \\
            \addlinespace
            & \multirow{2}{*}{gpt-3.5-turbo}
                & Precision & 0.6979 & 0.6800 & 0.4000 & 0.1923 & 0.7377 \\
            & & Recall      & 0.2199 & 0.1885 & 0.0291 & 0.0714 & 0.1637 \\
        \midrule
        \multirow{4}{*}{\emph{Q\&A}}
            & \multirow{2}{*}{gpt-4o-mini}
                & Precision & 0.6442 & 0.5789 & 0.0288 & 0.1338 & 0.7475 \\
            & & Recall      & 0.1263 & 0.1120 & 0.0146 & 0.0885 & 0.1452 \\
            \addlinespace
            & \multirow{2}{*}{gpt-3.5-turbo}
                & Precision & 0.4799 & 0.2799 & 0.0083 & 0.0561 & 0.6756 \\
            & & Recall      & 0.0956 & 0.0740 & 0.0101 & 0.0787 & 0.1340 \\
        \bottomrule
    \end{tabular}
\end{table*}

\begin{figure}
    \centering
    \includegraphics[width=\linewidth]{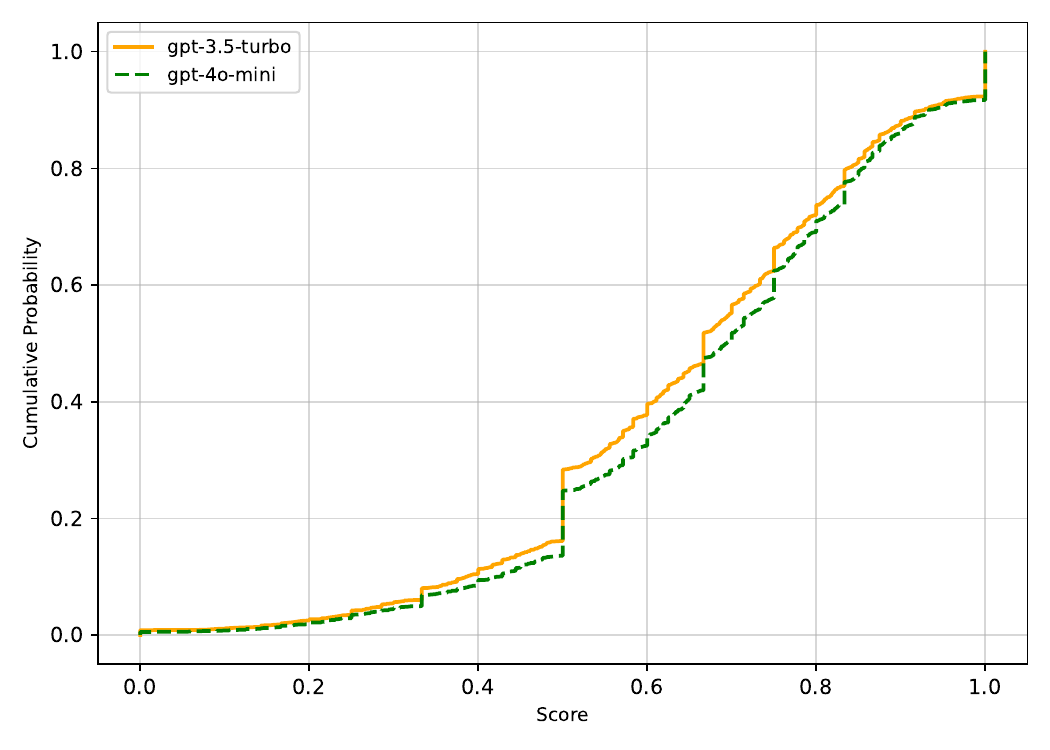}
    \caption{CDF of the frequency-based entity similarity between the source texts and their summaries. Lower CDF curves (toward the bottom right) indicate better similarity, as more samples achieve higher similarity values.}
    \label{fig:freq-cdf-overall}
\end{figure}

\begin{figure}
    \centering
    \subfloat[gpt-3.5-turbo\label{fig:loss-cdf-35}]{
        \includegraphics[width=\linewidth]{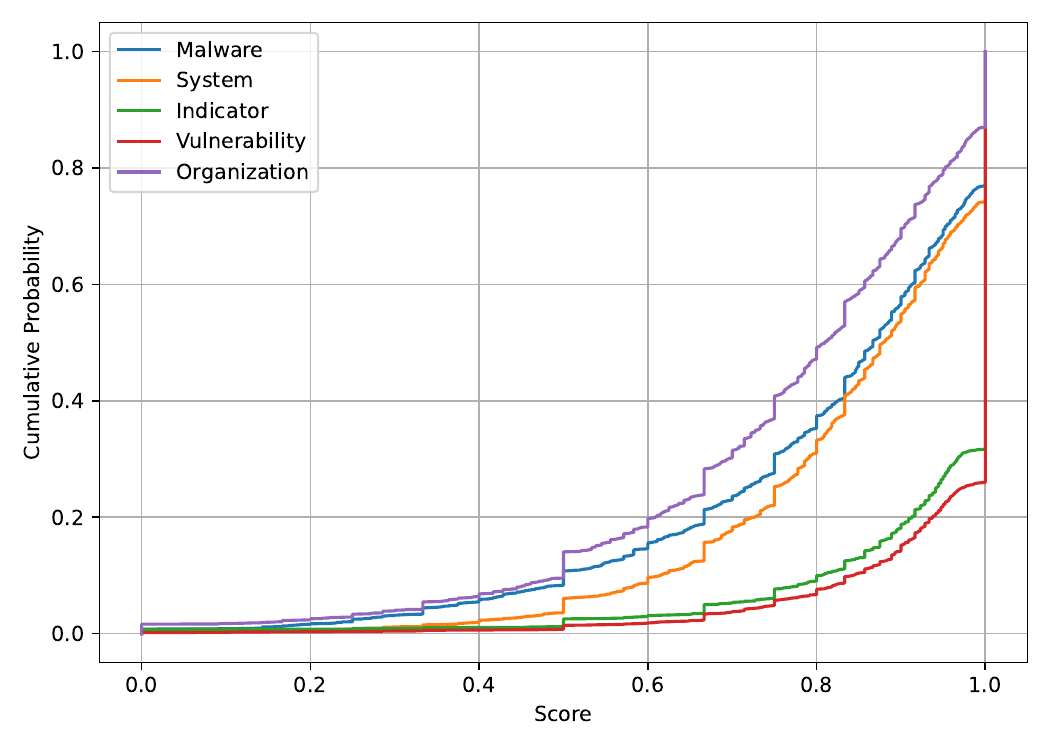}
    }
    \hfill
    \subfloat[gpt-4o-mini\label{fig:loss-cdf-4o}]{
        \includegraphics[width=\linewidth]{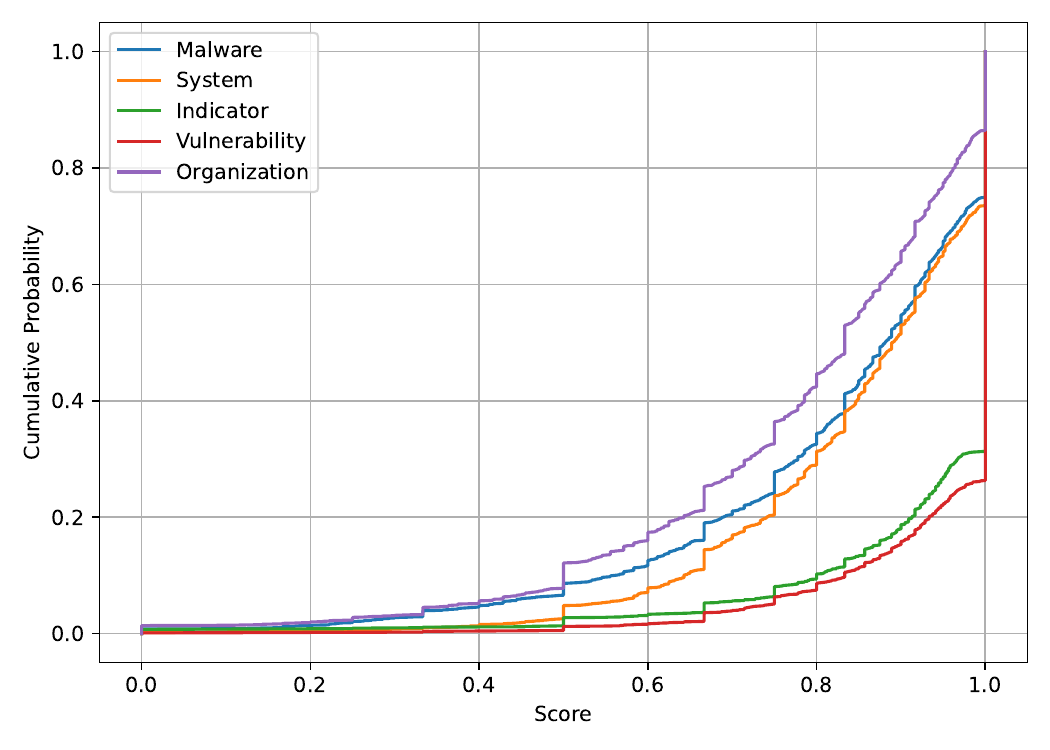}
    }
    \hfill
    
    \caption{CDF of frequency-based entity similarity between the source texts and their summaries \emph{for each label}. Lower CDF curves (toward the bottom right) indicate better similarity, as more samples achieve higher similarity values.}
    \label{fig:entity-freq-loss-comparison}
\end{figure}

We also performed a label-wise analysis of answer matching similarities by computing the average precision and recall for each entity type, as summarized in Table~\ref{tab:AnswerMatchingLabelwise}.  
The results show that the \texttt{Malware}, \texttt{System}, and \texttt{Organization} labels achieved precision scores of approximately 0.7 or higher, with \texttt{gpt-4o-mini} consistently outperforming \texttt{gpt-3.5-turbo}.  
In contrast, \texttt{Indicator} and \texttt{Vulnerability} had substantially lower precision (below 0.5 and 0.2, respectively) and recall (both below 0.1).  
This disparity likely stems from the nature of these two labels, which often involve non-linguistic entities such as IP addresses, port numbers, and CVEs (Common Vulnerabilities and Exposures), elements that are structurally diverse and therefore less reliably preserved in abstractive summarization.

However, high precision alone does not rule out the risk of hallucination. As our case analyses reveal, summaries may still exhibit factual inconsistencies or entity-level omissions even when precision is perfect. This highlights the need for multi-perspective evaluation frameworks that consider both precision and recall to capture the full spectrum of hallucination phenomena.

\noindent\textbf{Frequency-based Entity Similarity.}
We computed frequency-based entity similarity between the 4,940 articles and the summaries generated by \texttt{gpt-4o-mini} and \texttt{gpt-3.5-turbo}. Figure~\ref{fig:freq-cdf-overall} shows the overall results.  Most documents (over 60\%) achieve similarity scores above 0.6, indicating that the summaries largely preserve the relative distributions of entity types from the original texts. This suggests that both models demonstrate a strong capacity to maintain semantic fidelity at the entity level. Notably, \texttt{gpt-4o-mini} yields slightly higher similarity scores than \texttt{gpt-3.5-turbo}, reflecting more consistent preservation of entity-type proportions. This performance gap is particularly pronounced for semantically ambiguous categories such as \textit{Organization}, where fine-grained contextual reasoning is required. These findings imply that \texttt{gpt-4o-mini} produces summaries that more effectively capture the structural balance of entity information present in the source documents.

We also computed frequency-based entity similarity \emph{for each label}, as shown in Figure~\ref{fig:entity-freq-loss-comparison}. Both \texttt{gpt-4o-mini} and \texttt{gpt-3.5-turbo} exhibit broadly similar trends in their CDF curves. Notable differences emerge upon closer inspection of individual entity labels. Specifically, for the \texttt{Malware} and \texttt{System} labels, the CDF curves of both models show that over 50\% of the documents achieve a similarity score above 0.8. This indicates that the information gap for these labels is relatively small, suggesting that the models are generally successful in capturing and aligning these entity types between the source documents and the generated summaries.

In contrast, the CDF curves for \texttt{Vulnerability} and \texttt{Indicator} rise sharply at a score of 1.0, where about 70--75\% of the documents score exactly 1.0. However, this is mainly because CyNER rarely extracts these entities (Section~\ref{subsec:ner_reliability}): in many documents, neither the source nor the summary contains an extracted entity of that label, resulting in artificially high similarity despite a lack of actual alignment.

These distributional differences do not directly indicate hallucinations. However, we conjecture that large gaps in label distributions may be associated with hallucinations, as LLM-generated summaries can significantly deviate from the original articles in terms of entity composition. Given that correctly identifying entity labels is critical in the CTI domain, such discrepancies may serve as indirect evidence of hallucination. Therefore, analysts can use these metrics to conduct coarse-grained searches for hallucination candidates across large summary corpora.

\begin{figure}
    \centering
    \subfloat[Precision\label{fig:precision_qa}]{
        \includegraphics[width=\linewidth]{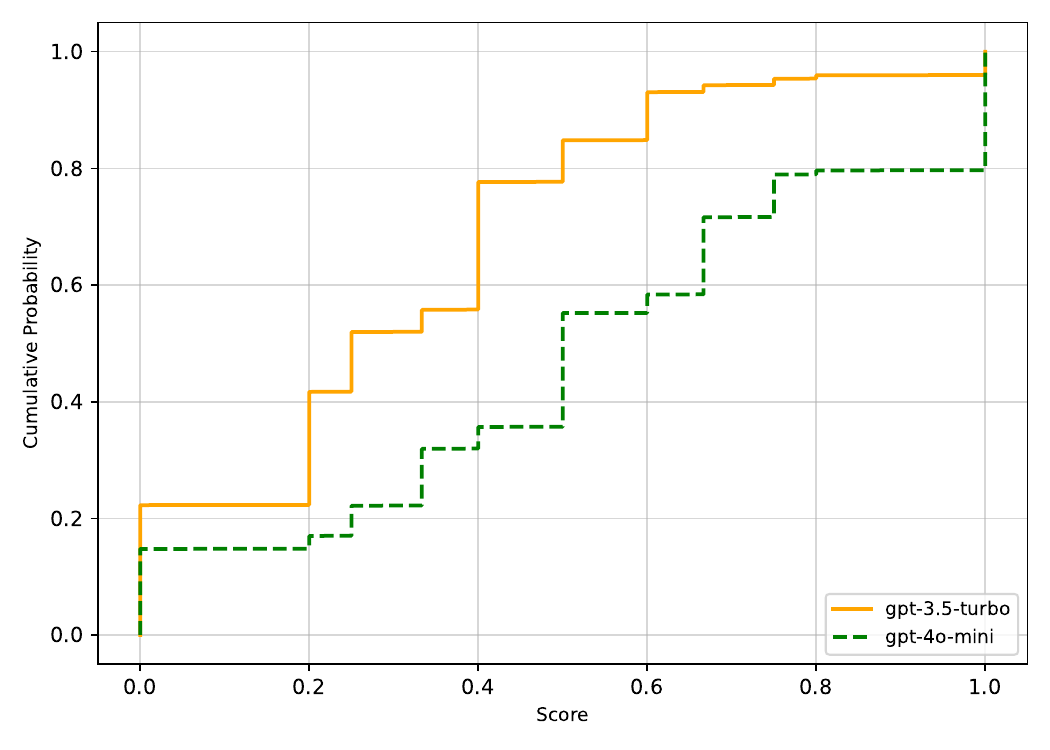}
    }
    \hfill
    \subfloat[Recall\label{fig:recall_qa}]{
        \includegraphics[width=\linewidth]{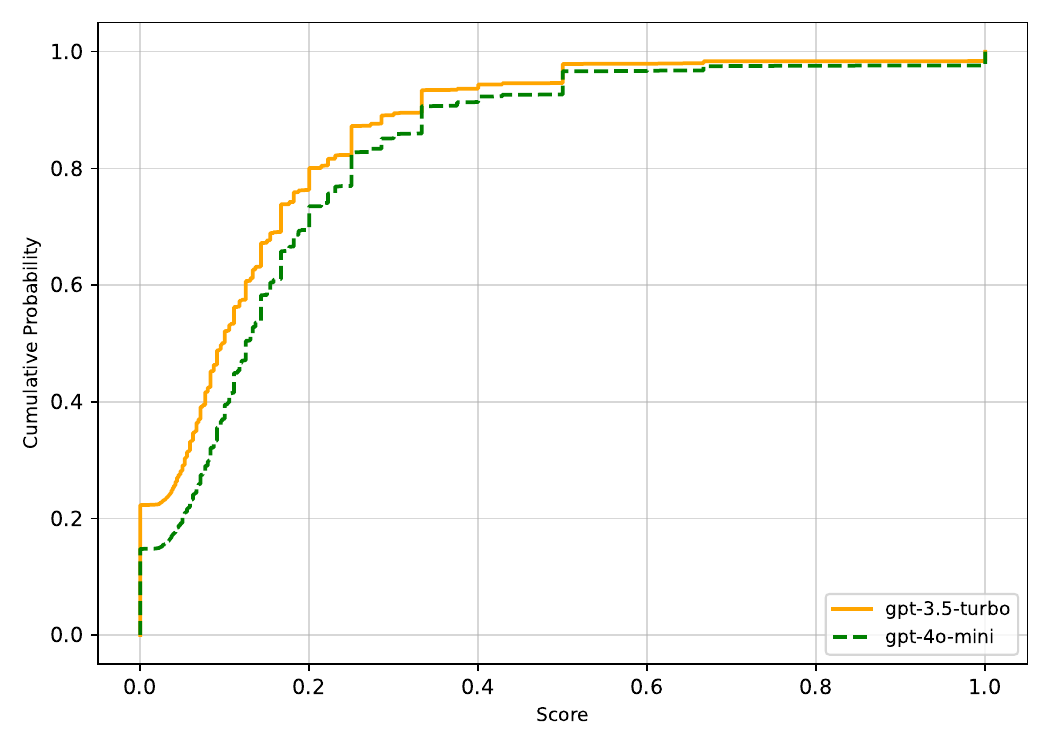}
    }
    \caption{CDF of answer matching similarities in the Q\&A task. Lower CDF curves (toward the bottom right) indicate better similarity, as more samples achieve higher precision or recall.}
    \label{fig:qa-prec-recall-cdf}
\end{figure}

\subsubsection{Q\&A}

To evaluate hallucinations in the Q\&A task, we assessed whether the label assigned to each entity in the LLM response matched its corresponding label in the original document. A mismatch in labeling between the source document and the LLM output indicates an information gap, which we consider particularly critical, because such discrepancies may significantly affect the reliability of downstream applications.

\noindent\textbf{Answer Matching Similarity.}
Figure~\ref{fig:qa-prec-recall-cdf} presents the distributions of precision and recall scores across all documents, based on whether the entities extracted by the LLM for each of the five predefined labels match the corresponding label assignments in the source documents. We observe that, for both precision and recall, at least 70\% of the documents score below 0.5, suggesting a broad deviation in label consistency. Notably, the recall distribution is skewed toward lower values, with nearly 80\% of the documents showing recall scores below 0.2, indicating frequent failures to retrieve all relevant entities.
Part of this gap stems from the extractor rather than the model: CyNER truncates spans and misses indicators sufficiently often for correct answers to be scored as mismatches (Section~\ref{subsec:ner_reliability}).
Despite these challenges, we find it noteworthy that the \texttt{gpt-4o-mini} model consistently outperforms \texttt{gpt-3.5-turbo} in both precision and recall distributions, suggesting that its stronger overall performance also translates into more accurate label-consistent entity extraction, which is a critical factor in reducing hallucination under this evaluation metric.

We further analyze the answer matching similarity for each label, as shown in Table~\ref{tab:AnswerMatchingLabelwise}. Among the five entity labels, \texttt{Malware}, \texttt{System}, and \texttt{Organization}, which represent semantically distinct and uniquely identifiable concepts, achieve relatively high precision. In contrast, \texttt{Indicator} and \texttt{Vulnerability} show relatively lower precision. For the same document set, \texttt{gpt-4o-mini} demonstrates higher precision than \texttt{gpt-3.5-turbo}, even for the challenging \texttt{Indicator} and \texttt{Vulnerability} labels. This suggests that the limitations associated with non-linguistic entity characteristics may be mitigated as model capabilities improve, highlighting a promising direction for future advancements.

\subsection{Open-weight Model Evaluation}
\label{appendix:open-weight}

\begin{table}[t]
    \footnotesize
    \centering
    \caption{Similarity between source articles and summaries generated by three LLMs using the 500-article subset. Cosine similarity is the mean across all articles; higher values indicate better performance for both metrics. Llama-3-8B-Instruct summaries are evaluated after removing a boilerplate opening line.}
    \label{tab:ow-summ-similarity}
    \resizebox{\columnwidth}{!}{
    \begin{tabular}{ l c c c }
        \toprule
        \multirow{2}{*}{\textbf{Model}} & \multirow{2}{*}{\makecell{\textbf{Cosine}\\\textbf{Similarity}}} & \multicolumn{2}{c}{\textbf{Answer Matching Similarity}} \\
        \cmidrule(lr){3-4}
        & & \textbf{Precision} & \textbf{Recall} \\
        \midrule
        Llama-3-8B-Instruct & 0.8659 & 0.7126 & 0.1547 \\
        gpt-3.5-turbo & 0.8847 & 0.6917 & 0.1766 \\
        gpt-4o-mini & \textbf{0.8911} & \textbf{0.7219} & \textbf{0.1831} \\
        \bottomrule
    \end{tabular}
    }
\end{table}

\begin{table}[t]
    \footnotesize
    \centering
    \caption{Q\&A answer matching precision by label on the 500-article subset, after format normalization, with overall precision and recall.}
    \label{tab:ow-qa-match}
    \resizebox{\columnwidth}{!}{
    \begin{tabular}{ l r r r }
        \toprule
        \textbf{Label} & \makecell{\textbf{Llama-3-8B-}\\\textbf{Instruct}} & \textbf{gpt-3.5-turbo} & \textbf{gpt-4o-mini} \\
        \midrule
        \texttt{Malware} & 0.3002 & 0.3489 & \textbf{0.4527} \\
        \texttt{System} & 0.2617 & 0.2353 & \textbf{0.3642} \\
        \texttt{Indicator} & \textbf{0.0142} & 0.0101 & 0.0101 \\
        \texttt{Vulnerability} & 0.0183 & 0.0304 & \textbf{0.0503} \\
        \texttt{Organization} & 0.6227 & 0.6491 & \textbf{0.7022} \\
        \midrule
        \emph{Overall} Precision & 0.3936 & 0.3073 & \textbf{0.5221} \\
        \emph{Overall} Recall & 0.7343 & 0.7890 & \textbf{0.8511} \\
        \bottomrule
    \end{tabular}
    }
\end{table}

To examine whether \ourtool{} and its findings generalize to an open-weight model, we also ran \texttt{Meta-Llama-3-8B-Instruct} through the same pipeline on the 500-article subset summarized in Table~\ref{tab:scrapdata}. The two GPT models were re-scored on the same 500 articles; hence, every comparison below is paired at the document level. The open-weight model was served in-process through Hugging Face \texttt{transformers} in bfloat16 on a single GPU with left-padded batching (batch size 8) and greedy decoding, matching the configuration reported in Appendix~\ref{appendix:api-config}. Generating outputs for both tasks on the subset (1,000 generations in total) took 10.5 minutes. For comparison, the OpenAI runs were inexpensive at list prices: generating outputs for both tasks for the full 4,940-article corpus consumed roughly 9.2 million input and 0.5 million output tokens per model, amounting to about \$1.69 for \texttt{gpt-4o-mini} and \$5.40 for \texttt{gpt-3.5-turbo} (Appendix~\ref{appendix:api-config}, Table~\ref{tab:api-config}).

For abstractive summarization (Table~\ref{tab:ow-summ-similarity}), the open-weight model exceeds \texttt{gpt-3.5-turbo} in answer matching precision and comes within 0.01 of \texttt{gpt-4o-mini}. However, it has the lowest recall of the three: its summaries contain fewer entities from the source, while those they include are comparatively accurate. It also ranks last on cosine similarity, trailing \texttt{gpt-4o-mini} by 0.0252. For Q\&A (Table~\ref{tab:ow-qa-match}), the open-weight model ranks between the two commercial GPT models, trailing \texttt{gpt-4o-mini} by 0.1285 precision points but leading \texttt{gpt-3.5-turbo} by 0.0863, although it has the lowest recall of the three.

Two conclusions follow. First, \ourtool{} is vendor-agnostic in practice, not merely by construction: an open-weight model can be evaluated using the same pipeline and metrics without changing the metric definitions. Second, entity-level fidelity in this domain does not follow a vendor-based ranking. An 8B open-weight model outperforms \texttt{gpt-3.5-turbo}, suggesting that hallucination propensity here tracks model generation and capacity rather than provider identity.

\subsection{Second-Source Evaluation}
\label{subsec:second-source}

\begin{table*}
    \footnotesize
    \centering
    \caption{Label-wise answer matching similarities for the 500-article sample from \textit{The Hacker News}, comparable to Table~\ref{tab:AnswerMatchingLabelwise}.}
    \label{tab:thehackernews}
    \begin{tabular}{l l l c c c c c}
        \toprule
        \textbf{Task} & \textbf{Model} & \textbf{Metric} & \textbf{Malware} & \textbf{System} & \textbf{Indicator} & \textbf{Vulnerability} & \textbf{Organization} \\
        \midrule
        \multirow{4}{*}{Abstractive Summarization}
            & \multirow{2}{*}{gpt-4o-mini}
                & Precision & 0.6549 & 0.6939 & 0.4595 & 0.2032 & 0.8158 \\
            & & Recall    & 0.1760 & 0.1537 & 0.0405 & 0.0777 & 0.2322 \\
        \addlinespace
            & \multirow{2}{*}{gpt-3.5-turbo}
                & Precision & 0.6851 & 0.7167 & 0.4571 & 0.2434 & 0.8058 \\
            & & Recall    & 0.1812 & 0.1630 & 0.0381 & 0.0701 & 0.1974 \\
        \midrule
        \multirow{4}{*}{Q\&A}
            & \multirow{2}{*}{gpt-4o-mini}
                & Precision & 0.5214 & 0.5619 & 0.0430 & 0.1840 & 0.7723 \\
            & & Recall    & 0.1001 & 0.1048 & 0.0191 & 0.0813 & 0.1662 \\
        \addlinespace
            & \multirow{2}{*}{gpt-3.5-turbo}
                & Precision & 0.3591 & 0.3116 & 0.0274 & 0.0935 & 0.6929 \\
            & & Recall    & 0.0800 & 0.0655 & 0.0310 & 0.0660 & 0.1529 \\
        \bottomrule
    \end{tabular}
\end{table*}

All results reported so far come from a single article source, Threatpost, and could therefore reflect its editorial style rather than CTI reporting in general. To assess transferability, we repeated the measurement on a second source excluded from all stages of the experimental setup: 500 articles from The Hacker News, selected using label-stratified sampling. Both models were run through the same pipeline using the same decoding configuration, and the reference entities were extracted utilizing the same fine-tuned CyNER model used for Threatpost, ensuring that both corpora were evaluated using the same measurement procedure. Table~\ref{tab:thehackernews} reports answer matching similarity in the same format as Table~\ref{tab:AnswerMatchingLabelwise}.

We make two observations. First, the metrics exhibit the same characteristic pattern for The Hacker News articles. The label profile is preserved, with \texttt{Organization} and \texttt{Malware} scoring highest and \texttt{Indicator} lowest in both corpora, while precision remains far above recall in summarization for the reason discussed in Section~\ref{subsec:quantitave}. The values shift only modestly, indicating that the framework does not degrade when applied to text outside its development corpus. Second, the Q\&A comparison also holds: \texttt{gpt-4o-mini} leads \texttt{gpt-3.5-turbo} in precision across all five labels and in recall across four of the five, consistent with Table~\ref{tab:AnswerMatchingLabelwise}. In contrast, the summarization comparison does not hold as consistently. The two models differ by no more than 0.04 on most labels in that task, and in this smaller sample \texttt{gpt-3.5-turbo} achieves slightly higher precision on three of the five labels.

\subsection{Human Validation}
\label{subsec:human_validation}

To confirm that the scores reported by \ourtool{} correspond to hallucinations as perceived by human readers, we conducted a blinded annotation study. We recruited 10 annotators whose academic majors were cybersecurity or artificial intelligence. We sampled 100 \texttt{gpt-4o-mini} outputs from our 4,940-document corpus using a stratified design: documents were grouped into quartiles based on summarization precision and, independently, Q\&A precision, and approximately six documents were selected from each of the resulting 16 strata. This ensured that the annotated set spanned the full range of metric scores, including outputs that scored high on one task and low on the other, rather than being concentrated in the modal region of the distribution.

Each annotator was shown the source article together with both the generated summary
and the Q\&A answers and assigned a single rating to the pair: \emph{faithful}, \emph{partially
supported}, or \emph{hallucinated}; the rating is therefore not attributable
to a single task\footnote{Note that because cosine similarity applies mainly to summarization while answer matching covers both tasks, this joint rating does not reveal which task triggered a \emph{hallucinated} verdict, and the per-metric comparison below should be interpreted with caution.}. Annotators saw neither the model identity nor any metric scores. They received only a one-line definition for each label\footnote{Annotators were instructed to judge only against the source article, with paraphrasing and reasonable generalization allowed. The labels were defined as follows: \emph{faithful}, the summary and all five answers are fully supported by the article; \emph{partially supported}, some entity is absent or misattributed, but the overall meaning holds; \emph{hallucinated}, the output introduces a fact, number, or actor not in the article.}, with no worked examples and no calibration round; we kept the instructions minimal to avoid biasing the ratings with our own notion of hallucination. Every output was independently rated by three of the ten annotators under a rotating assignment,
yielding 300 ratings. Table~\ref{tab:human_cases} shows representative cases.

\begin{figure*}[t]
    \centering
    \subfloat[Cosine similarity\label{fig:hv-cos}]{
        \includegraphics[width=0.32\textwidth]{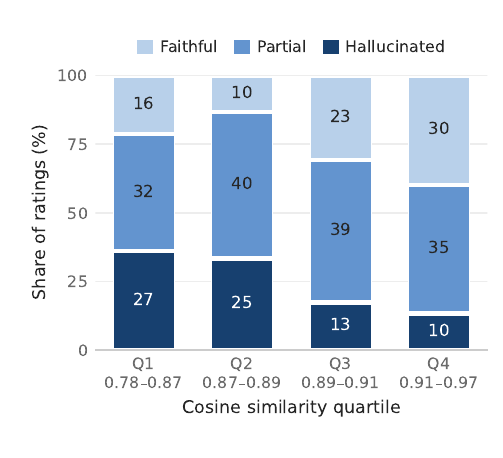}}
    \hfill
    \subfloat[Answer matching similarity\label{fig:hv-ams}]{
        \includegraphics[width=0.32\textwidth]{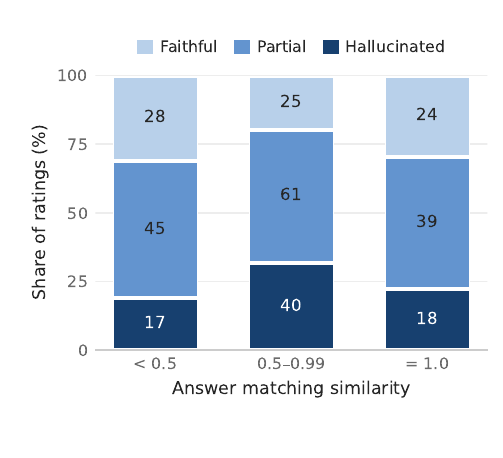}}
    \hfill
    \subfloat[Frequency-based entity similarity\label{fig:hv-fes}]{
        \includegraphics[width=0.32\textwidth]{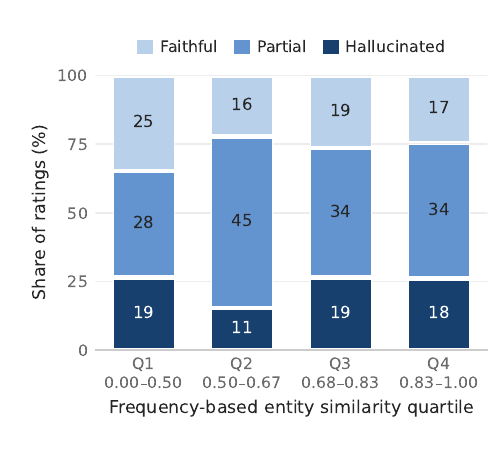}}
    \caption{Human ratings of the 100 annotated outputs, grouped by metric score.
    Each output was rated by three annotators, so each group contains three ratings per output.
    Answer matching similarity is grouped by value rather than by quartile because 27 of the
    99 outputs have a score of exactly 1.0; frequency-based entity similarity is defined for 95 outputs.}
    \label{fig:human_validation}
\end{figure*}

\begin{table}
    \footnotesize
    \caption{Examples of human validation case studies. Scores are \ourtool{} metrics expressed in percentages (CS: cosine similarity, AMS: answer matching similarity (precision), FES: frequency-based entity similarity). Each output was independently rated by three annotators as F (faithful), P (partially supported), or H (hallucinated), with the three ratings listed in order.}
    \centering
    \label{tab:human_cases}
    \begin{tabular}{ c c c c c }
        \toprule
        \textbf{Doc} & \textbf{CS} & \textbf{AMS} & \textbf{FES} & \textbf{Human Rating} \\
        \midrule
        \#11   & 79.3 & 33.3 & 25.0 & H H H \\
        \#2078 & 82.1 & 33.3 & 78.6 & H P H \\
        \#3916 & 95.2 & 100.0 & 75.0 & F F F \\
        \bottomrule
    \end{tabular}
\end{table}

Figure~\ref{fig:human_validation} illustrates the human ratings for each metric score
range. For cosine similarity, the proportion of \emph{hallucinated} ratings decreases
monotonically across quartiles, from 27 of 75 ratings in the lowest quartile to 10 of 75 in
the highest, while \emph{faithful} ratings increase. In contrast, the
two entity-level metrics show no such pattern, which we attribute to two properties of these
metrics. First, they capture only named entities, whereas a large share of the problems cited
by the annotators involve added claims or interpretations, such as recommendations or generalizations
unsupported by the source, which no NER-based comparison can detect. Second, answer
matching similarity has minimal resolution at the top of its range: 27 of the 99 outputs score
exactly 1.0, and the ratings within that block (24 \emph{faithful}, 39 \emph{partially
supported}, 18 \emph{hallucinated}) match the distribution of the sample as a whole. Thus,
perfect entity precision carries no information about whether human readers regard the output
as faithful.

Inter-annotator agreement was Krippendorff's $\alpha = 0.29$, indicating fair rather than substantial agreement. Of the 100 outputs, 34 were rated unanimously, ratings for 62 differed only between adjacent categories, and only four received both a \emph{faithful} and a \emph{hallucinated} rating; merging \emph{faithful} and \emph{partially supported} increased $\alpha$ to 0.43. The annotators thus agreed on clear cases but drew the \emph{partially supported} boundary differently, which we attribute to the subjectivity of deciding whether a paraphrased, generalized, or missing entity still counts as fully supported, particularly for annotators without CTI analysis experience, and to our minimal instructions. Future studies should provide worked examples and a calibration round. Given these results, we position our metrics as screening signals rather than calibrated detectors, because establishing an operating point would require fitting and evaluating on disjoint labeled sets, which a sample of 100 outputs cannot support. The entity-based metrics indicate which entity types diverge, while cosine similarity helps rank LLM outputs for human review.

\subsection{Comparison with Existing Hallucination Detectors}
\label{subsec:baselines}

FActScore~\citep{min2023factscore} and QAGS~\citep{wang2020asking} are established reference points for faithfulness evaluation. Both decompose generated text into units: atomic facts in the first case and question--answer pairs in the second, and check those units against the source using auxiliary models. However, these approaches use evaluation units that differ from those of \ourtool{}, which compares the source and generated texts using sentence embeddings and typed CTI entities. Therefore, the extent of agreement between the two approaches remains unclear. To measure this agreement, we reimplemented both detectors and ran them alongside the three proposed metrics on a 200-document subset comprising the 100 documents used in the human validation study (Section~\ref{subsec:human_validation}) and another 100 randomly selected documents.

\begin{table*}
    \footnotesize
    \centering
    \caption{Three documents for which FActScore reports full support, although at least two of the three annotators rated the output \emph{partially supported} or \emph{hallucinated}. Human ratings show the verdict of each of three independent annotators, each of whom was shown the complete article: F (faithful), P (partially supported), and H (hallucinated). The last column indicates the most common reason among the non-faithful ratings. Cos.\ sim.\ denotes cosine similarity, and AM prec.\ denotes answer matching precision.}
    \label{tab:baseline-cases}
    \begin{tabular}{l c c c c c l}
        \toprule
        & \multicolumn{2}{c}{\textbf{Baselines}} & \multicolumn{2}{c}{\textbf{\ourtool{}}} & & \\
        \cmidrule(lr){2-3} \cmidrule(lr){4-5}
        \textbf{Doc} & \textbf{FActScore} & \textbf{QAGS} & \textbf{Cos.\ sim.} & \textbf{AM prec.} & \textbf{Human Rating} & \textbf{Main reason given} \\
        \midrule
        \#4042 & 1.000 & 0.546 & 0.923 & \textbf{0.000} & P P H & Unsupported elaboration \\
        \#3313 & 1.000 & 0.709 & 0.929 & \textbf{0.250} & F P P & Wrong entity type \\
        \#4087 & 1.000 & 0.660 & 0.940 & 1.000 & P P P & Unsupported elaboration \\
        \bottomrule
    \end{tabular}
\end{table*}

We observed that both baselines scored almost every summary as fully faithful, and that neither varied with the proposed metrics. FActScore assigned a score of exactly 1.000 to 182 of the 200 summaries and judged only 13 of the 1,844 extracted atomic claims unsupported. QAGS remained close to its mean of 0.706 throughout. This saturation stems from the task itself, not from our reimplementation. Each summary is generated from the same article against which its claims are subsequently checked, whereas the biography-generation task for which FActScore reported 58 percent provided the model with no source text. Using the original prompt of Min \emph{et al.}~\citep{min2023factscore} resulted in minimal change: the number of perfect scores increased to 189 of 200.

Aggregate counts alone cannot show which method is correct; hence, we inspected the individual disagreements. Table~\ref{tab:baseline-cases} lists documents that FActScore rates as fully supported but the annotators do not, one for each cause we observed. The first cause is unsupported elaboration: FActScore only asks whether a claim is supported, so a plausible generalization that the source neither states nor contradicts still passes. QAGS responds to such additions, because an assertion that appears only in the summary produces a question the source cannot answer. However, this occurs for only 32 of the 1,000 questions in the subset, and one unanswerable question out of five shifts the document score by less than one standard deviation relative to scores for summaries that no annotator questioned.

The second cause is an incorrect entity type. In document~\#3313, the distributed denial-of-service (DDoS) campaign is mentioned in the source article, as confirmed by both baseline methods. However, the model incorrectly reported it as an indicator. In contrast, our answer matching similarity detects both cases, while cosine similarity does not, as expected for a document-level semantic measure when an error involves a single entity. The two metrics therefore address different questions: whether the generated text follows from its source, and which threat entities are preserved in their respective roles. On this corpus, the first question is almost always answered affirmatively, whereas the second continues to distinguish documents. We therefore consider our metrics complementary to judge-based detectors rather than replacements and acknowledge that determining which metric best matches expert judgment would require a larger and more reliable validation set than the one available here.

\subsection{Reliability of the CyNER Ground Truth}
\label{subsec:ner_reliability}

All entity-level measurements in \ourtool{} rely on CyNER output; hence, NER errors can be indistinguishable from LLM hallucinations unless the ground truth is itself validated. To verify the ground truth, one of our authors  manually audited CyNER's entity recognition using a $\sqrt{N}$ sample of the corpus: 70 of the 4,940 Threatpost articles, stratified into quartiles based on summary precision and Q\&A precision for gpt-4o-mini, so that the sample spanned the range of measured performance rather than concentrating on easy documents. 

Within each article, up to three predicted entities were classified as \emph{Exact}, \emph{Partial}, or \emph{Spurious}. For the latter two categories, we classified NER errors as \emph{Boundary} (a correct entity that is truncated or fragmented), \emph{Label} (a genuine entity assigned the wrong type), and \emph{Not Entity} (tagged text that is not an entity). This procedure yielded 204 verified entity judgments. Because entities were sampled within documents, each judgment is weighted by its inverse inclusion probability, so the 204 judgments represent all 1,036 entities CyNER predicted in those articles. Every article was also read in full to identify entities the model failed to predict.

\begin{table*}[!t]
    \footnotesize
    \centering
    \caption{Manual validation of the CyNER ground truth on 70 articles.
    $n$ is the number of judged predictions; precision and recall are
    inverse-probability weighted and provide estimates for all 1,036
    predictions in those articles. A prediction is correct only if both its span and
    label are correct; rejections are attributed to \emph{Boundary}
    (truncated span, e.g. \emph{vil} for \emph{REvil}), \emph{Label} (wrong
    type, e.g. \emph{Samsung} tagged \texttt{System}), or \emph{Not Entity} (not
    an entity, e.g. \emph{Top Kill} tagged \texttt{Malware}). \emph{Missed}
    counts entities CyNER did not predict.}
    \label{tab:ner-manual}
    \begin{tabular}{l c c c c c c c}
        \toprule
        & & & & \multicolumn{3}{c}{\textbf{NER Errors}} & \\
        \cmidrule(lr){5-7}
        \textbf{Label} & $n$ & \textbf{Precision} & \textbf{Recall} & \textbf{Boundary} & \textbf{Label} & \textbf{Not Entity} & \textbf{Missed} \\
        \midrule
        Malware       & 56 & 0.626 & 0.976 & 9 & 5 & 7 & 5 \\
        System        & 50 & 0.733 & 0.929 & 6 & 3 & 4 & 15 \\
        Indicator     & 12 & 0.968 & 0.771 & 0 & 2 & 0 & 18 \\
        Vulnerability & 3$^{\dagger}$ & 0.000 & 0.000 & 2 & 1 & 0 & 22 \\
        Organization  & 83 & 0.760 & 0.969 & 7 & 5 & 8 & 9 \\
        \midrule
        \textbf{Overall} & \textbf{204} & \textbf{0.710} & \textbf{0.914} & \textbf{24} & \textbf{16} & \textbf{19} & \textbf{69} \\
        \bottomrule
    \end{tabular}
    \\[2pt]
    \footnotesize{$^{\dagger}$Only three \texttt{Vulnerability} spans were
    predicted in the sample, so the precision and recall reported for this label
    are indicative only.}
\end{table*}

Table~\ref{tab:ner-manual} summarizes the manual validation results. Overall precision was 0.710 and recall was 0.914, compared with the 74.69\% precision and 77.84\% recall that CyNER achieved on its held-out evaluation set (Section~\ref{subsec:ex_setup}). Thus, precision was comparable to, and recall substantially higher than, those obtained on the held-out evaluation set, indicating no systematic degradation on the corpus to which the model is actually applied. Boundary truncation accounted for 13.5\% of predictions and incorrect labels for 7.0\%, while only 8.5\% were not named entities: 91.5\% of what CyNER predicted was an entity that genuinely occurred in the article, and most of the
shortfall in precision was because of surface-form errors rather than invented entities.
Because answer matching compares surface forms exactly, a
truncated span such as \emph{vil} for \emph{REvil} or \emph{Over} for
\emph{GameOver} is scored as a mismatch even though the underlying entity was
recognized.

More importantly, CyNER predicted only 3 \texttt{Vulnerability} spans
across the 70 articles, and none was exactly correct, whereas the annotator identified 22 vulnerability entities that CyNER missed. The low precision
reported for this label in Table~\ref{tab:AnswerMatchingLabelwise} therefore
reflects a genuine weakness of the extractor rather than a property of the LLM
outputs, and results for \texttt{Vulnerability} should be interpreted as a lower bound
on model fidelity. To assess the reliability of these judgments, another author independently re-judged all predictions in 20 of the 70 articles (58 of the 204 judgments). The two annotators agreed on 49 of the 58 predictions (Cohen's $\kappa = 0.73$). This subset contained no \texttt{Vulnerability} predictions, and the 22 missed \texttt{Vulnerability} entities were identified by a single annotator, so the agreement measure does not cover that label. The per-label samples for \texttt{Vulnerability} and \texttt{Indicator} also remain small.

\subsection{Separating Extractor and Model Error}
\label{subsec:decomp}

CyNER errors reduce the answer matching scores in two ways: an entity that CyNER misses
in the generated text cannot be matched even if the LLM reproduced it, and a
span that CyNER truncates or mislabels on either side counts as a mismatch. Assuming
these errors are independent of what the LLM generated, the measured score is approximately equal to the
true score multiplied by a correction factor $c$, and the true score is at most
$\text{measured}/c$. We derive $c$ from Table~\ref{tab:ner-manual}: for
summarization, where CyNER processes both texts, $c$ is the product of its precision and
recall; for question answering, where the answers are compared directly, $c$ is
its recall alone. Because this assumes that every extractor error causes a mismatch,
the result is an upper bound on what the extractor can explain.

\begin{table*}[!t]
    \footnotesize
    \centering
    \caption{Answer matching similarity before and after accounting for the contribution of
    CyNER errors (measured $\rightarrow$
    $\min(1, \text{measured}/c)$). The difference between the two values represents the maximum contribution 
    of extractor errors; the difference between the second value and 1.0 represents the
    minimum contribution attributable to the LLM. \texttt{Vulnerability} is
    omitted because CyNER did not recover any of its spans in the audit ($c = 0$).}
    \label{tab:extractor-decomp}
    \begin{tabular}{l l l c c c c}
        \toprule
        \textbf{Task} & \textbf{Model} & \textbf{Metric} & \textbf{Malware} & \textbf{System} & \textbf{Indicator} & \textbf{Organization} \\
        \midrule
        \multicolumn{3}{l}{Correction factor $c$ (Summarization: $P \times R$ of CyNER)} & 0.611 & 0.681 & 0.746 & 0.736 \\
        \multicolumn{3}{l}{Correction factor $c$ (Q\&A: $R$ of CyNER)} & 0.976 & 0.929 & 0.771 & 0.969 \\
        \midrule
        \multirow{4}{*}{\shortstack[l]{Abstractive\\Summarization}}
            & \multirow{2}{*}{gpt-4o-mini}
                & Precision & 0.740 $\rightarrow$ 1.000 & 0.719 $\rightarrow$ 1.000 & 0.498 $\rightarrow$ 0.667 & 0.764 $\rightarrow$ 1.000 \\
            & & Recall    & 0.224 $\rightarrow$ 0.367 & 0.192 $\rightarrow$ 0.283 & 0.032 $\rightarrow$ 0.042 & 0.179 $\rightarrow$ 0.243 \\
            \addlinespace
            & \multirow{2}{*}{gpt-3.5-turbo}
                & Precision & 0.698 $\rightarrow$ 1.000 & 0.680 $\rightarrow$ 0.999 & 0.400 $\rightarrow$ 0.536 & 0.738 $\rightarrow$ 1.000 \\
            & & Recall    & 0.220 $\rightarrow$ 0.360 & 0.189 $\rightarrow$ 0.277 & 0.029 $\rightarrow$ 0.039 & 0.164 $\rightarrow$ 0.222 \\
        \midrule
        \multirow{4}{*}{\shortstack[l]{Question \&\\Answering}}
            & \multirow{2}{*}{gpt-4o-mini}
                & Precision & 0.644 $\rightarrow$ 0.660 & 0.579 $\rightarrow$ 0.623 & 0.029 $\rightarrow$ 0.037 & 0.748 $\rightarrow$ 0.771 \\
            & & Recall    & 0.126 $\rightarrow$ 0.129 & 0.112 $\rightarrow$ 0.121 & 0.015 $\rightarrow$ 0.019 & 0.145 $\rightarrow$ 0.150 \\
            \addlinespace
            & \multirow{2}{*}{gpt-3.5-turbo}
                & Precision & 0.480 $\rightarrow$ 0.492 & 0.280 $\rightarrow$ 0.301 & 0.008 $\rightarrow$ 0.011 & 0.676 $\rightarrow$ 0.697 \\
            & & Recall    & 0.096 $\rightarrow$ 0.098 & 0.074 $\rightarrow$ 0.080 & 0.010 $\rightarrow$ 0.013 & 0.134 $\rightarrow$ 0.138 \\
        \bottomrule
    \end{tabular}
\end{table*}

Table~\ref{tab:extractor-decomp} reveals three findings (figures below are
for \texttt{gpt-4o-mini}; \texttt{gpt-3.5-turbo} shows the same pattern). First,
recall is barely affected. In summarization, \texttt{Indicator} recall increases
only from 0.032 to 0.042, and recall for \texttt{Malware}, \texttt{System}, and
\texttt{Organization} increases from 0.224, 0.192, and 0.179 to 0.367, 0.283, and 0.243, respectively;
even after the correction, most source entities are absent from the summary.
This holds even if CyNER's \texttt{Indicator} recall were as low as 0.5, which
would increase the bound only to 0.065. Second, question answering precision is
also robust: no label changes by more than 0.05 (e.g., \texttt{Malware} from 0.644 to
0.660 and \texttt{Indicator} from 0.029 to 0.037), so extractor errors cannot explain
the precision loss in question answering.

Third, summarization precision for \texttt{Malware},
\texttt{System}, and \texttt{Organization} cannot be separated into extractor and LLM contributions: the measured
values (0.740, 0.719, 0.764) divided by $c$ (0.611, 0.681, 0.736) already
exceed 1.0, which means that CyNER's own error rate on these labels is as
large as the measured loss. Only for \texttt{Indicator} does the bound remain below 1.0 (0.667),
indicating a loss of at least 0.333 that the extractor cannot explain. For these three labels, the precision loss in summarization therefore cannot be taken
as evidence of LLM hallucination, although the comparison between the two models remains
unaffected because both are measured with the same extractor.

\subsection{Case Studies}
\label{subsec:case_studies}

In this section, we qualitatively examine articles that received low scores in the quantitative evaluation, providing case-based evidence of hallucinations.

\begin{table*}
    \footnotesize
    \centering
    \caption{A real hallucination example in abstractive summarization using \texttt{gpt-3.5-turbo}, in which an LLM-generated summary exhibits low cosine similarity (79.35\%) with the original text.}
    \resizebox{\textwidth}{!}{
    \begin{tabular}{p{3cm} p{13cm}}
        \toprule
        \textbf{Original Article} & 
        [...] \textit{\textcolor{LimeGreen}{Wikileaks' Web site has been the target of denial of service attacks}}, which began even before the group published the leaked diplomatic cables as part of a campaign it dubbed ``Cablegate.'' While the source of those attacks isn't clear, a hacker using the handle th3j35t3r claimed responsibility for it, citing the need to retaliate for the leak. More recently, \textit{\textcolor{LimeGreen}{Wikileaks has enlisted the services of hundreds of mirror sites}} to host the cablegate documents in a coordinated effort to keep them from being driven from the Internet. [...] \\
        \midrule
        \textbf{Summarized Text} (\textit{hallucinated}) & 
        Paypal permanently restricted Wikileaks' account due to policy violation, sparking online feud with hacker-activists. Anonymous planned DDoS attack on Paypal in response. \textit{\textcolor{red}{Wikileaks faced ongoing denial of service attacks, with mirror sites hosting leaked documents.}} PayPal and Amazon denied services to Wikileaks, as U.S. Attorney General authorized actions against the organization. \\
        \midrule
        \textbf{Reason} & 
        The summary implies that the \textbf{DDoS attacks were associated with or originated through the mirror sites.} However, the original text clearly states that \textbf{mirror sites were a defensive measure in response to DDoS attacks.} \\
        \bottomrule
    \end{tabular}
    }
    \label{tab:hallucination_summarization_example}
\end{table*}

\begin{table*}
    \centering
    \footnotesize
    \caption{Real hallucination examples in Q\&A tasks using the \texttt{gpt-3.5-turbo} model.}
    \resizebox{\textwidth}{!}{
    \begin{tabular}{p{5cm} p{3cm} p{2cm} p{2cm} p{5cm}}
    \toprule
        \textbf{Original Article} & \textbf{Hallucinated Entity} & \textbf{Ground-Truth} (\textbf{CyNER}) & \textbf{LLM-generated Answer} (\textbf{gpt-3.5-turbo}) & \textbf{Reason} \\
    \midrule
        And while it's in vogue to refer to \textcolor{blue}{\textbf{REvil}} and DarkSide as ``franchise models'' which supply Ransomware-as-a-Service, it is important to remember that the franchisees are effectively freelance cybercriminals. & REvil & \textcolor{LimeGreen}{Malware} & \textcolor{red}{Organization} & REvil is a MALWARE in this article, not an ORGANIZATION. \\
    \midrule
        Now researchers from \textcolor{blue}{\textbf{Blackberry}} have connected the dots, painting a picture of a malware that has evolved five times in twelve months. & Blackberry & \textcolor{LimeGreen}{Organization} & \textcolor{red}{System} & Blackberry is correctly labeled as an Organization, not a System, as it acts as a sighting group. \\
    \midrule
        \textcolor{blue}{\textbf{Log4j}}, ransomware, cloud vulnerabilities, phishing: Cyber threats are manifold. They all pale, however, in comparison to the security black holes that walk around on two legs. & Log4j & \textcolor{LimeGreen}{Not Identified} & \textcolor{red}{Malware} & Log4j is a Java-based library and should be classified as a System, not Malware. \\
    \midrule
        APT29, a\/k\/a Cozy Bear, has used \textcolor{blue}{\textbf{Tor}} and a technique called domain fronting in order to secure backdoor access to targets for nearly two years running. & Tor & \textcolor{LimeGreen}{System} & \textcolor{red}{Indicator} & Tor is a System as a browser, not an INDICATOR. \\
    \bottomrule
    \end{tabular}
    }
    \label{tab:QAAnswerMatchingExamples}
\end{table*}

\noindent\textbf{Hallucinated Causality in Abstractive Summarization.}
Section~\ref{subsec:quantitave} characterizes the distribution of each metric rather than establishing a detection threshold. Here, we select cases from the lowest-scoring summaries for qualitative inspection. For instance, we observed a clear discrepancy in a text pair with a cosine similarity of 79.35\%. As illustrated in Table~\ref{tab:hallucination_summarization_example}, the original document states:
\begin{quote}
\emph{(..) WikiLeaks is hosted through mirror sites in response to DDoS attacks (..)}
\end{quote}
However, the LLM-generated summary incorrectly claims:
\begin{quote}
\emph{(..) DDoS attacks are being carried out through hosting mirror sites. (..)}
\end{quote}
The summary erroneously implies that the mirror sites were responsible for launching the DDoS attacks, whereas the original text clearly describes them as a countermeasure. This misrepresentation exemplifies hallucinated causality in the context of security incidents, potentially leading to confusion among security professionals.

\noindent\textbf{Hallucinated Entities in Q\&A.} Mislabeling of key entities by LLMs can significantly undermine the accuracy of NER tasks. In many cases, entity labels extracted from LLM-generated responses deviate from the ground truth, revealing challenges in extracting reliable information from generated content. We observed instances in which an LLM incorrectly classified entities from the source text. For example, \texttt{Log4j}, a Java-based logging library, was misclassified as \texttt{Malware} instead of its correct category, \texttt{System}. Similarly, \texttt{REvil}, a well-known piece of malicious software, was often mislabeled as an \texttt{Organization}. Such misclassifications, illustrated in Table~\ref{tab:QAAnswerMatchingExamples}, contribute to reduced precision in entity recognition.

Furthermore, we observed instances in which the LLM generated incorrect responses due to misaligned \emph{in-context learning}. In-context learning is a paradigm in which language models acquire task behavior from a few examples provided in the prompt, without parameter updates~\citep{NEURIPS2020_1457c0d6,dong2024surveyincontextlearning}. While LLMs typically infer the expected output format from such demonstrations or system instructions, they may sometimes misinterpret the prompt and produce responses that deviate from the intended structure. As illustrated in Appendix~\ref{appendix:prompts}, we used triple backticks (\textasciigrave\textasciigrave\textasciigrave) in our prompts to explicitly highlight the target span for entity extraction. However, in several cases, the model incorrectly incorporated the backticks into the generated output, suggesting that it learned an unintended formatting pattern from the context.
For example, instead of returning a clean answer such as \texttt{REvil}, the model produced verbose outputs such as \texttt{Your Answer for Question 1: REvil} or injected escape sequences, as in \texttt{Answer 1: REvil\textbackslash nAnswer 2: Servers}, despite the prompt clearly indicating the desired format.

\section{Discussion}
\label{sec:discussion}

\noindent\textbf{Gap between Hallucination Metrics.} Given the lack of a universally accepted definition of hallucination in the context of language models, there is also no single, standardized metric for its quantitative evaluation~\citep{ji2023survey}. This limitation underscores the importance of employing multiple metrics when analyzing hallucination. For example, \emph{cosine similarity} serves as a semantic-level measure by embedding both the source and generated texts into dense vector representations and computing their angular similarity. In contrast, \emph{answer matching similarity} evaluates hallucinations from a lexical perspective by comparing entities extracted via NER, providing a more direct measure of the consistency of surface-level information. Meanwhile, \emph{frequency-based entity similarity} assesses the alignment of entity label distributions between the source and generated texts, aiming to detect information shifts at the document-level structural granularity.  Because each metric targets a different aspect of hallucination, \emph{semantic, lexical, or structural}, the resulting similarity scores may diverge for the same input pair.

To test whether the differences between the models were statistically significant, we compared \texttt{gpt-4o-mini} and \texttt{gpt-3.5-turbo} on each overall score using a paired bootstrap over documents (10,000 resamples). Although the metrics target different aspects of hallucination, they agree on which model stays closer to the source: as shown in Table~\ref{tab:model-sig}, \texttt{gpt-4o-mini} scores higher on every metric tested, and each difference exceeds sampling noise. The gap is concentrated in question answering, where a model must commit to a single entity per label, whereas both models can reproduce much of the source wording in summarization.

\begin{table}[h]
    \footnotesize
    \centering
    \caption{Paired bootstrap comparison of \texttt{gpt-4o-mini} with
    \texttt{gpt-3.5-turbo} (overall score, 10,000 resamples). AS = Abstractive
    Summarization, QA = Question answering, Cossim = cosine similarity,
    AM-prec = answer matching precision, and Freq-sim = frequency-based entity
    similarity. All differences are statistically significant at
    $p<0.0001$, but the margin is substantial only in question answering. Positive $\Delta$ values indicate that \texttt{gpt-4o-mini} is closer to the source. $n$ varies because each metric is computed only for documents for which both models produced a valid output and the metric is defined.}
    \label{tab:model-sig}
    \resizebox{\linewidth}{!}{
    \begin{tabular}{l l c c}
        \toprule
        \textbf{Task} & \textbf{Metric} & \textbf{$n$} & \textbf{$\Delta$ (4o-mini $-$ 3.5-turbo), 95\% confidence interval} \\
        \midrule
        QA & AM-prec & 4,843 & +0.2228 \, [0.2151, 0.2307] \\
        \midrule
        AS & AM-prec & 4,940 & +0.0310 \, [0.0210, 0.0400] \\
        AS & Cossim  & 4,939 & +0.0070 \, [0.0063, 0.0077] \\
        AS & Freq-sim & 4,490 & +0.0157 \, [0.0100, 0.0212] \\
        \bottomrule
    \end{tabular}
    }
\end{table}

\noindent\textbf{Practical Implications of Misclassification.} The critical impact of our findings is most evident in how LLM misclassifications directly compromise essential cybersecurity applications. For instance, a single misclassified IoC can contaminate the entire intelligence feed in commercial threat intelligence platforms\footnote{For example, S2W QUAXAR (\url{https://s2w.inc/en/product/pd02}) and AhnLab Threat Intelligence Platform (\url{https://www.ahnlab.com/en/product/threat-intelligence}).}, triggering automated defenses that result in business-disrupting service outages. This corrupted data can then mislead Security Orchestration, Automation, and Response (SOAR) platforms\footnote{For example, Splunk SOAR (\url{https://www.splunk.com/en_us/products/splunk-security-orchestration-and-automation.html}) and Elastic Security for SOAR (\url{https://www.elastic.co/security/soar}).}, causing them to execute flawed playbooks or ignore critical threats, thereby defeating the very purpose of automation. Ultimately, entity misclassification within these core applications systematically undermines an organization's security posture and erodes analyst trust, highlighting the urgent need for domain-specific safeguards.

\noindent\textbf{Scope of the Study.}
Our study focuses exclusively on intrinsic hallucinations, in which LLM outputs contradict the given source text. We acknowledge, however, that extrinsic hallucinations, cases in which the model generates information that cannot be verified against the source, can also have serious consequences in cybersecurity scenarios. For instance, an LLM-generated answer may introduce a plausible-sounding but nonexistent CVE identifier or misattribute an attack to an unrelated threat actor, potentially misleading downstream analysis even when the source article itself contains no such claim. We deliberately scoped our analysis to intrinsic hallucinations for two reasons. 

First, our measurement framework compares LLM outputs against the original source articles, providing a clear and reproducible reference for detecting intrinsic inconsistencies. Assessing extrinsic hallucinations, by contrast, would require an external ground-truth corpus of cybersecurity facts that is rarely available or up to date in the rapidly evolving CTI landscape. Second, extrinsic hallucinations largely originate from gaps between the model's parametric knowledge and external facts and can therefore be substantially mitigated through knowledge-augmentation techniques such as Retrieval-Augmented Generation (RAG)~\citep{lewis2020retrieval}. Intrinsic hallucinations, in contrast, stem from errors in the model's reasoning over the provided context and are less amenable to such external remedies, making them a more fundamental reliability challenge that merits dedicated study. We further note that intrinsic and extrinsic hallucinations may co-occur within a single output. Disentangling them in mixed cases remains an open problem. We leave a systematic analysis of extrinsic hallucinations and their interaction with intrinsic ones in CTI for future work.

\section{Limitations and Future Work}
\label{sec:limitations}

\noindent\textbf{Thresholds for Similarity Scores.} Although similarity-based metrics offer a useful signal, they are neither necessary nor sufficient for detecting hallucinations and, importantly, do not admit a universal threshold. Let $s_e(x)$ denote the similarity score of an LLM-generated text $x$ and $\tau_e$ the hallucination threshold for entity type $e$. A hallucination may be flagged when $s_e(x) < \tau_e$. However, choosing $\tau_e$ is inherently application-specific and data-dependent. In security-sensitive contexts, the cost of accepting incorrect facts may far outweigh the cost of additional review, motivating thresholds tuned for higher precision (at the expense of recall). Conversely, in broad threat hunting, recall (i.e., avoiding missed detections) may be prioritized. In principle, $\tau_e^\ast$ can be selected on a small labeled development set by minimizing a cost-sensitive risk, but such calibration requires ground truth that is often scarce in CTI.

\noindent\textbf{Complementary Nature of Proposed Metrics.} No single metric perfectly captures all dimensions of hallucination, a challenge widely acknowledged in the NLP literature~\cite{ji2023survey}. Cosine similarity, for instance, may assign high scores to summaries that paraphrase or reorder content while still introducing factual errors, as demonstrated in our case study (Section~\ref{subsec:case_studies}). This is precisely why HalluVision adopts a multi-metric design: cosine similarity detects semantic drift at the document level, answer matching similarity captures entity-level factual inconsistency, and frequency-based entity similarity identifies structural divergence in entity distributions. Together, these metrics provide complementary signals that are more robust than any single measure. We recommend that practitioners combine all three scores rather than relying on any one metric in isolation and interpret low scores as indicators warranting further human review rather than definitive hallucination labels. We note that this recommendation concerns interpreting the three measurements jointly and should not be interpreted as a claim that collapsing them into a single number improves detection; the combined score defined in Appendix~\ref{appendix:combined} is a way of expressing a single decision, and its comparison with the individual components is left to future work.

\noindent\textbf{Ground-truth Dataset.}
A potential limitation of our study lies in how ground truth is established. Our methodology relies on a fine-tuned CyNER model, which achieves an F1-score of 76.23\%, representing a reasonable level of performance given that fine-tuning was conducted on only 50 manually annotated MITRE articles, though admittedly not state-of-the-art. We argue that this does not fundamentally compromise our analysis for two reasons. First, the NER model is applied symmetrically to both the source text and the LLM-generated output, so extraction noise affects both sides comparably, and the relative trends we report (e.g., \texttt{gpt-4o-mini} consistently outperforming \texttt{gpt-3.5-turbo}, or large gaps across entity labels) remain meaningful. 

Second, CyNER is one instantiation of the NER component, not a fixed dependency. Entity ambiguity also remains a challenge. For example, \emph{Log4j} may be labeled as \texttt{System} when referring to the software library, or as \texttt{Vulnerability} when denoting a specific security flaw. The generated ground truth may not always distinguish such contexts consistently. A Q\&A mismatch may therefore reflect an incorrect answer, an extraction error, or an ambiguous label. Section~\ref{subsec:decomp} bounds only the extraction error, so a mismatch indicates disagreement with the extracted reference rather than a verified hallucination.
This reflects the specific NER model employed rather than the framework itself.
The NER module is a pluggable component by design and can be readily replaced with stronger cybersecurity NER models without altering the rest of the pipeline; the precision of \ourtool{}'s evaluation will improve accordingly as such models emerge. Thus, the key contribution of this work lies not in the performance of any single component but in introducing a general and extensible framework for automated LLM hallucination analysis in the CTI domain.

\noindent\textbf{Evaluation of a Limited Number of LLMs.}
This study focuses on a small number of models, including OpenAI's GPT models and Meta's
Llama-3-8B-Instruct, to evaluate hallucinations in the context of CTI. We adopted
this approach primarily because OpenAI provides accessible APIs that facilitate
large-scale interaction with LLMs, enabling efficient collection of model-generated
texts, while an open-weight model can be served locally under fully controlled
decoding settings. However, this focus naturally limits the generalizability of our
findings, as other LLMs, such as Claude, Gemini, and CTI-tuned
open-source models, differ in architecture, training data, and alignment strategies,
which may lead to variations in hallucination behavior. Future work should extend this
analysis to include such diverse models to provide a more comprehensive comparison
across LLM families.

\noindent\textbf{Temporal Relevance of Results.}
This study was initiated when \texttt{gpt-3.5} was the most widely accessible and cost-effective API model, enabling large-scale experimentation.
Although newer models such as \texttt{gpt-4o} have since been released with significantly improved fidelity, the hallucination patterns observed at the time of the study remain relevant for understanding earlier-generation model limitations and for benchmarking future models. Moreover, our evaluation framework is model-agnostic and remains applicable to assessing hallucination in current and emerging LLMs, particularly in high-stakes domains such as CTI.

\begin{table*}[t]
    \centering
    \caption{Comparison of related studies with our work.}
    \footnotesize
    \resizebox{\textwidth}{!}{
    \begin{tabular}{l >{\raggedright\arraybackslash}m{5cm} m{2.5cm} m{2.5cm} m{2.5cm}}
    \toprule
         \textbf{Work} & \textbf{Scope} & \textbf{CTI Entity Extraction} & \textbf{Hallucination Analysis} & \textbf{Cybersecurity-specific} \\ 
    \midrule
         Liao et al.~\citep{liao2016acing} & IoC extraction from cybersecurity articles & \CIRCLE & \Circle & \CIRCLE \\ \midrule
         TTPDrill~\citep{husari2017ttpdrill} & CTI ontology extraction from unstructured text & \CIRCLE & \Circle & \CIRCLE \\ \midrule
         LLM-TIKG~\citep{HU2024103999} & LLM-based CTI knowledge graph construction from unstructured text & \CIRCLE & \Circle & \CIRCLE \\ \midrule
         GapFinder~\citep{jo2020gapfinder} & Information-gap analysis in cybersecurity articles & \CIRCLE & \Circle & \CIRCLE \\ \midrule
         Jiang et al.~\citep{jiang2021evaluating} & Information-gap analysis in cybersecurity databases & \CIRCLE & \Circle & \CIRCLE \\ \midrule
         Vulcan~\citep{jo2022vulcan} & CTI entity extraction from unstructured text & \CIRCLE & \Circle & \CIRCLE \\ \midrule
         HaluEval~\citep{li2023halueval} & Hallucination benchmark for large language models & \Circle & \CIRCLE & \Circle \\ \midrule
         ReEval~\citep{yu2023automatic} & Hallucination assessment via adversarial examples & \Circle & \CIRCLE & \Circle \\ \midrule
         KGV~\citep{wu2026kgv} & CTI credibility assessment via LLM-driven claim verification & \CIRCLE & \CIRCLE~(LLM-judge) & \CIRCLE \\ \midrule
         CTIBench~\citep{alam2024ctibench} & Benchmark of LLM knowledge and reasoning on CTI tasks & \Circle & \Circle & \CIRCLE \\ \midrule
         Mezzi et al.~\citep{mezzi2025unreliable} & Consistency and confidence of LLMs on CTI reports & \Circle & \CIRCLE~(consistency) & \CIRCLE \\ \midrule
         \textbf{\ourtool{} (this work)} & Hallucination analysis in the cybersecurity domain & \CIRCLE & \CIRCLE~(statistical \& NER-based) & \CIRCLE \\ 
    \bottomrule
    \end{tabular}
    }
    \label{tab:comparison}
\end{table*}

\section{Related Work}
\label{sec:related}

A substantial body of research has explored both CTI and hallucinations in LLMs.

\noindent\textbf{Cyber Threat Intelligence.} Several studies have proposed methodologies for mining security information from diverse sources to support CTI. Husari \emph{et al.}~\citep{husari2017ttpdrill} introduced TTPDrill, which automatically generates a CTI ontology from unstructured text, while Hu \emph{et al.}~\citep{HU2024103999} more recently proposed LLM-TIKG, which leverages LLMs to construct CTI knowledge graphs from OSINT. Liao \emph{et al.}~\citep{liao2016acing} proposed a system that automatically analyzes IoCs from open-source articles and blogs to collect CTI information effectively. Jo \emph{et al.}~\citep{jo2022vulcan} presented a novel CTI system capable of extracting various forms of CTI data and their semantic relationships from unstructured textual sources. More recently, Alam \emph{et al.}~\citep{alam2024ctibench} introduced CTIBench, a benchmark that evaluates the CTI knowledge and reasoning abilities of LLMs across several task types. 

\noindent\textbf{Information Gap in CTI.} The trustworthiness of information provided by CTI is crucial, as security experts depend on it to develop effective countermeasures. However, the accuracy of CTI information has been questioned in several studies.  
Jo \emph{et al.}~\citep{jo2020gapfinder} proposed a framework called GapFinder, which extracts structured relations from unstructured text to identify inconsistencies in publicly available security articles. They showed that information gaps exist between articles from different sources, such as malware discovery reports, and that these gaps can significantly hinder malware analysis. Jiang \emph{et al.}~\citep{jiang2021evaluating} investigated inconsistencies across cybersecurity databases, including the Common Vulnerabilities and Exposures (CVE) and the National Vulnerability Database (NVD). Their findings revealed that critical attributes, such as the Common Vulnerability Scoring System (CVSS), often vary between databases, potentially causing confusion in risk assessments.  
In this context, our work focuses on verifying hallucinations by comparing ground-truth content (i.e., the original text) with LLM-generated text, aiming to assess information inconsistencies that may arise when CTI is supported by LLMs.

\noindent\textbf{Hallucination Assessment for LLMs.}  
Hallucination refers to the generation of inaccurate or fabricated information by  LLMs and is widely recognized as a critical challenge in the NLP community~\citep{ji2023survey}. To mitigate this issue, recent LLMs have been aligned through instruction tuning and human feedback. However, hallucinations persist. Consequently, numerous studies have been conducted to assess and characterize hallucination behaviors in LLMs.  
Yu \emph{et al.}~\citep{yu2023automatic} proposed ReEval, a framework that evaluates hallucinations in retrieval-augmented LLMs through transferable adversarial attacks. Li \emph{et al.}~\citep{li2023halueval} introduced HaluEval, a benchmark designed to assess LLMs’ ability to recognize hallucinated content. Closer to our setting, Mezzi \emph{et al.}~\citep{mezzi2025unreliable} evaluated three LLMs on 350 threat intelligence reports and found their outputs inconsistent and overconfident, with few-shot learning and fine-tuning offering only partial remedies. Their study measures the stability and confidence of LLM answers, whereas we measure the agreement between LLM outputs and the source article at the semantic and entity levels. 
In contrast to these works, our study aims to evaluate hallucinations in the cybersecurity domain, where inaccurate or misleading information can have more severe consequences than in general-purpose applications.

\noindent\textbf{Comparison with previous studies.}
Table~\ref{tab:comparison} summarizes the differences between prior studies and our work. Prior studies have focused on traditional approaches to CTI entity extraction. Recently, hallucinations have raised concerns about the use of LLMs for domain-specific tasks, as errors such as misidentifying CTI entities or misattributing a tactic to the wrong threat actor can lead to severe security risks. However, existing analyses of hallucination are largely limited to general, non-specialized settings. In contrast, our study specifically targets the cybersecurity domain and systematically evaluates hallucinations generated by LLMs on CTI texts. 

Closest to our work, Wu \emph{et al.}~\citep{wu2026kgv} proposed KGV, which integrates LLMs with knowledge graphs to assess the credibility of CTI reports through LLM-driven claim verification. Although KGV shares our motivation of detecting unreliable information in the CTI domain, it differs from our work in two key respects: (i) KGV evaluates the credibility of CTI articles themselves, whereas we evaluate hallucinations in LLM-generated outputs against the source articles; and (ii) KGV relies on LLM-as-judge verification, which may inherit the very hallucination tendencies it seeks to detect, while \ourtool{} adopts lightweight statistical and NER-based metrics that are transparent and reproducible without requiring additional LLM judgments. Sood \emph{et al.}~\citep{SOOD2025110307} presented a conceptual taxonomy\footnote{Note that, unlike the system- and methodology-oriented works summarized in Table~\ref{tab:comparison}, Sood et al.~\cite{SOOD2025110307} is a conceptual survey rather than an empirical framework, and is therefore discussed separately here rather than included in the comparison table.} of hallucinations specific to cybersecurity LLMs, categorizing them into contextual, factual, attributional, ambiguous, and logical types, and discussed corresponding mitigation strategies. Although their work offers useful conceptual categories for reasoning about hallucinations in security contexts, it remains descriptive and does not provide an empirical methodology for measuring such hallucinations on real-world data, which our work aims to address.

\section{Conclusion}
\label{sec:conclusion}

This study measured how LLMs hallucinate when they summarize CTI articles and answer entity-level questions about them. \ourtool{} compared 4,940 \emph{Threatpost} articles with the outputs of \texttt{gpt-3.5-turbo} and \texttt{gpt-4o-mini} using cosine similarity, answer matching similarity, and frequency-based entity similarity. The analysis was extended to an open-weight model and a second news source. The summaries are mostly accurate regarding the entities they include but omit most of the entities in the source: for \texttt{gpt-4o-mini}, answer matching precision averages 0.702, whereas recall averages only 0.182, and the low recall persists after correcting for extractor errors. The two OpenAI models differ mainly in question answering rather than in summarization. In a blinded study of 100 outputs rated by ten annotators, only cosine similarity tracks human judgment, while existing hallucination detectors such as FActScore and QAGS rate nearly every source-grounded summary as faithful, including outputs that the annotators judged partially supported or hallucinated.

These measurements have clear limitations. The entity extractor misses most \texttt{Vulnerability} entities, so results for that label are a lower bound on model fidelity, and the human ratings show only fair agreement. The metrics are therefore best used as screening signals to rank outputs for human review rather than as calibrated hallucination detectors. For practitioners, an LLM summary of a threat report should not be treated as a complete record of its indicators and affected systems, and entities from low-recall labels warrant manual checking before they enter a threat intelligence platform. Future work should replace the extractor with a stronger CTI-specific model, collect separate human labels for each task with calibrated guidelines, and extend the measurement to newer LLM families.

\section*{Acknowledgements}
This work was supported by the National Research Foundation of Korea (NRF) grant funded by the Korea government (MSIT) (No. RS-2026-25495427).

\appendix

\section{Prompt Templates}
\label{appendix:prompts}

\begin{tcolorbox}[title=Summarization Prompt,fonttitle=\bfseries,colback=gray!5,colframe=gray!50,breakable]
\# Main Role \\
- You're a famous NER Expert in CTI (Cyber Threat Intelligence) domain. Your task is to generate a short summary of a malware information obtained from popular malware blog posts, following the instructions below.\\

\# Main Instructions \\
- Summarize the content below in at least \textbf{50} words, and at most \textbf{70} words.\\
- You should answer using only the words and contents in the summary as much as possible.\\

\#\# Input blog text data \\
\texttt{\{original\_text\}}
\end{tcolorbox}

\begin{tcolorbox}[title=Question \& Answering Prompt,fonttitle=\bfseries,colback=gray!5,colframe=gray!50,breakable]
\# Main Role \\
- You're a famous NER Expert in CTI (Cyber Threat Intelligence) domain. Your task is to generate several questions that ask meaningful information about the content of a malware article retrieved from a popular malware blog post.\\

\# Main Instructions \\
- Answer questions from blog text data given as input. The questions are also specified below. Answer each question in a section enclosed by three backticks.\\
- Please note that these blogs are all related to CTI.\\
- All answers to the questions must be short answers. It would be great if you could answer the question with just one entity.\\
- If you determine that there is no appropriate answer to a question, you MUST print ``no answer''.\\

\#\# Input blog text data\\
\texttt{\{original\_text\}}\\

\# Questions\\
Question 1: What is the main MALWARE covered in this article?\\
Answer 1: \textasciigrave\textasciigrave\textasciigrave Your Answer for Question 1 \textasciigrave\textasciigrave\textasciigrave\\
Question 2: What is the main SYSTEM covered in this article? \\ 
Answer 2: \textasciigrave\textasciigrave\textasciigrave Your Answer for Question 2 \textasciigrave\textasciigrave\textasciigrave \\
Question 3:  What is the main INDICATOR covered in this article? \\
Answer 3: \textasciigrave\textasciigrave\textasciigrave Your Answer for Question 3 \textasciigrave\textasciigrave\textasciigrave \\
Question 4: What is the main VULNERABILITY covered in this article? \\
Answer 4: \textasciigrave\textasciigrave\textasciigrave Your Answer for Question 4 \textasciigrave\textasciigrave\textasciigrave \\
Question 5: What is the main ORGANIZATION covered in this article?\\
Answer 5: \textasciigrave\textasciigrave\textasciigrave Your Answer for Question 5 \textasciigrave\textasciigrave\textasciigrave \\
\end{tcolorbox}

\section{Worked Examples of the Metrics}
\label{appendix:examples}

This appendix illustrates how each of the three metrics defined in Section~\ref{subsec:hallu_anal} is computed.

\noindent\textbf{Cosine Similarity.}
As an illustrative example (based on Table~\ref{tab:hallucination_text_comparison}), the source and summary texts are encoded into 1,024-dimensional vectors:
\begin{align*}
X' &= [1.361,\, -2.186,\, 0.669,\, \dots,\, 0.263] \\
Y' &= [2.033,\, -1.821,\, 0.553,\, \dots,\, 0.4397]
\end{align*}
Only the first three and last components are shown for brevity. The resulting cosine similarity is
\[
C(X, Y) = \frac{X' \cdot Y'}{\|X'\| \|Y'\|} \approx 0.708
\]
This score indicates a moderate level of semantic similarity between the original document and the generated summary.

\noindent\textbf{Answer Matching Similarity.}
For example, consider the \texttt{Malware} label.  
Suppose the gold entities are \texttt{[LockBit, Dridex, Evil, Zeus]}, and the LLM-generated prediction is \texttt{[LockBit, Hades, Dridex]}.  
In this case:
\begin{itemize}
  \item TP: \texttt{[LockBit, Dridex]}
  \item FP: \texttt{[Hades]}
  \item FN: \texttt{[Evil, Zeus]}
\end{itemize}
Accordingly, the precision and recall are computed as
\[
\text{Precision} = \frac{|\text{TP}|}{|\text{TP}| + |\text{FP}|} = \frac{2}{2 + 1} = \frac{2}{3} \approx 0.667
\]
\[
\text{Recall} = \frac{|\text{TP}|}{|\text{TP}| + |\text{FN}|} = \frac{2}{2 + 2} = \frac{2}{4} = 0.5
\]

\noindent\textbf{Frequency-based Entity Similarity.}
Consider the example shown in Table~\ref{tab:hallucination_text_comparison}, in which the original article $X$ contains entity counts  
$c^X = [4, 5, 6, 0, 3]$ and the LLM-generated summary $Y$ yields $c^Y = [1, 3, 0, 0, 1]$, corresponding to the labels \texttt{System}, \texttt{Malware}, \texttt{Indicator}, \texttt{Vulnerability}, and \texttt{Organization}.  
After normalization, the resulting distributions are:
\begin{align*}
p^X &= [0.2222,\, 0.2778,\, 0.3333,\, 0.0000,\, 0.1667] \\
p^Y &= [0.2000,\, 0.6000,\, 0.0000,\, 0.0000,\, 0.2000]
\end{align*}
The $L_1$ distance between the two distributions is $0.7110$, leading to a frequency-based entity similarity score:
\[
\text{F}(X, Y) = 1 - \frac{1}{2} \cdot 0.7110 = 0.6445
\]
This indicates moderate alignment between the original and generated entity distributions, with noticeable divergence particularly in the \texttt{Malware} and \texttt{Indicator} labels.

\section{Combining the Three Metrics into a Single Score}
\label{appendix:combined}

Reporting three numbers per document makes it harder to compare models or corpora at
a glance, raising the question of whether they can be reduced to one.
Algorithm~\ref{alg:combined} standardizes each metric over the corpus, averages the three
with equal weights, and negates the average so that larger values indicate greater
divergence from the source. The weights are fixed a priori rather than fitted to labels,
so the rule introduces no tunable parameters. Equal weights are a neutral default, not a validated choice. Without labeled data, there is no principled basis for preferring these weights over alternative weighting schemes.

\begin{algorithm}
    \caption{Combined similarity $S$. CS, AMS, and FES denote cosine, answer
    matching, and frequency-based entity similarity. No step is fitted to hallucination
    labels.}
    \label{alg:combined}
    \small
    \begin{algorithmic}[1]
        \Require metric values $x_{i,j}$ for documents $i = 1,\dots,N$ and
                 $j \in \{\mathrm{CS}, \mathrm{AMS}, \mathrm{FES}\}$; all three are
                 similarity measures, so each is defined to decrease as the output
                 diverges from its source along that metric's dimension
        \Ensure one score $S_i$ per document; larger means greater divergence
        \For{$j \in \{\mathrm{CS}, \mathrm{AMS}, \mathrm{FES}\}$}
            \State $m_j \gets \mathrm{mean}_i\, x_{i,j}$, \quad
                   $s_j \gets \mathrm{sd}_i\, x_{i,j}$
            \State $z_{i,j} \gets (x_{i,j} - m_j) / s_j$ \textbf{for all} $i$
                   \Comment{z-score standardization}
        \EndFor
        \State $S_i \gets -\frac{1}{3}\sum_j z_{i,j}$ \textbf{for all} $i$
               \Comment{equal weights, no free parameters}
        \State \Return $S_i$
    \end{algorithmic}
\end{algorithm}

\begin{figure}
    \centering
    \includegraphics[width=\linewidth]{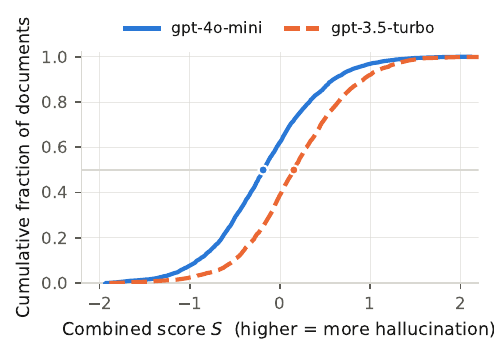}
    \caption{Combined score $S$ over the full corpus, representing the output of
    Algorithm~\ref{alg:combined} alone, with no labels or threshold. The three
    measurements are standardized jointly across both models, so the two curves share a
    common scale; the markers denote the medians ($-0.19$ for gpt-4o-mini and $+0.15$ for
    gpt-3.5-turbo).}
    \label{fig:combined_score}
\end{figure}

Applied to the full corpus, the rule preserves the ordering produced by the individual
measurements (Section~\ref{subsec:quantitave}). As shown in
Figure~\ref{fig:combined_score}, the mean combined score is centered at $+0.172$ for
\texttt{gpt-3.5-turbo} and $-0.167$ for \texttt{gpt-4o-mini}. The score separates the two
models, not individual documents. The curves overlap over most of their range, and their two
centers are separated by approximately half a pooled standard deviation. Nevertheless, \texttt{gpt-3.5-turbo}
achieves lower scores on 77.2\% of the 4{,}409 articles processed by both models and accounts for
70.0\% of the articles in the lowest-scoring decile of the corpus. The combined score therefore inherits its
ordering mainly from the entity-level measurement, which is where the two models differ
most once each metric is placed on a common scale, while the semantic and structural
measurements contribute only slightly to  the score. Equal weighting keeps that inheritance visible;
a fitted weighting would incorporate it into a coefficient.

Whether such a combination improves on the individual metrics is a separate question.
Addressing this question would require fitting and evaluating weights on disjoint labeled sets, which
the 100 annotated outputs of this study cannot support. We therefore report results for each
metric throughout the paper and leave a validated combination to future work.

\section{LLM API Configuration}
\label{appendix:api-config}

Both tasks (abstractive summarization and question answering) use identical settings. Table~\ref{tab:api-config} lists the configurations of all models used in this study, together with the token usage and costs associated with the OpenAI API calls.

\begin{table*}[t]
    \footnotesize
    \centering
    \caption{Generation configurations of the three LLMs used in this study, along with the token usage and list-price cost of the OpenAI runs. Main-corpus token counts were reconstructed using \texttt{tiktoken} from the stored prompts and outputs, because those runs predate usage logging; validation-run counts were obtained from the API usage logs and include the FActScore and QAGS judge calls. Prices are per million tokens at the time of the runs.}
    \label{tab:api-config}
    \begin{tabular}{ l l l }
        \toprule
        \textbf{Parameter} & \textbf{Value} & \textbf{Applies to} \\
        \midrule
        \multirow{3}{*}{Model snapshot}
            & \texttt{gpt-3.5-turbo-0125} & gpt-3.5-turbo \\
            & \texttt{gpt-4o-mini-2024-07-18} & gpt-4o-mini \\
            & \texttt{Meta-Llama-3-8B-Instruct} & Llama-3-8B-Instruct \\
        \midrule
        Temperature & 0.0 & All models \\
        Top-$p$ & 1.0 & All models \\
        Seed & 42 & OpenAI models \\
        Stop sequence & None & All models \\
        System message & None (user message only) & All models \\
        \midrule
        Max tokens & Unlimited & OpenAI models \\
        Max new tokens & 256 & Llama-3-8B-Instruct \\
        Decoding & Greedy & Llama-3-8B-Instruct \\
        Precision & bfloat16 & Llama-3-8B-Instruct \\
        Batch size & 8 (left-padded) & Llama-3-8B-Instruct \\
        \midrule
        \multirow{2}{*}{List price (input / output), per 1M tokens}
            & \$0.50 / \$1.50 & gpt-3.5-turbo \\
            & \$0.15 / \$0.60 & gpt-4o-mini \\
        \multirow{2}{*}{Cost, main corpus (4,940 articles, both tasks)}
            & \$5.40 & gpt-3.5-turbo \\
            & \$1.69 & gpt-4o-mini \\
        \bottomrule
    \end{tabular}
\end{table*}

\bibliographystyle{model1-num-names}

\bibliography{references.bib}

\newpage
\bio{}
\textbf{Seungho Song} is a researcher in the AIR (AI Research) Lab at Korea University. He received his B.S. degree from the School of Software at Kwangwoon University. His current research interests focus on trustworthy AI for large language models (LLMs), with a particular focus on model extraction attacks and LLM watermarking.
\endbio

\vspace{0.5in}

\bio{}
\textbf{Donggyu Lee} is a researcher in the Interactive 3D Media Lab at Korea University. He received his B.S. degree in artificial intelligence from the School of Software at Kwangwoon University. His current research interests include machine learning security, 3D vision, and real-time graphics.
\endbio

\vspace{0.5in}

\bio{}
\textbf{Jaehan Kim} is a Ph.D. candidate in the School of Electrical Engineering at KAIST. He received his M.S. and B.S. degrees in the School of Electrical Engineering at KAIST. His current research interests focus on cybersecurity, machine learning security, and large language models.
\endbio

\vspace{0.5in}

\bio{}
\textbf{Jiyoung Choi} was a researcher in the Cloud and Network Security (CNS) Lab at Kwangwoon University. She received her B.S. degree with a double major in the Department of Mathematics and the School of Software at Kwangwoon University. Her research interests include natural language processing and cyber threat intelligence.
\endbio

\vspace{0.5in}

\bio{}
\textbf{Jinwoo Kim} is an assistant professor in the Department of Computer Engineering at Chungbuk National University, Korea. He received his Ph.D. and M.S. degrees from KAIST and his B.S. degree from Chungnam National University. He was previously an assistant professor at Kwangwoon University. His research interests include security and privacy in software-defined networks, cloud systems, and virtual reality.
\endbio

\end{document}